\documentclass[10pt,journal,compsoc]{IEEEtran}
\usepackage{xcolor}
\usepackage[table]{xcolor}
\usepackage{color}
\usepackage{amsfonts}       % blackboard math symbols
\usepackage{amsmath}
\usepackage{graphicx}
\usepackage{multirow}
\usepackage{booktabs}
\usepackage{bbding}
\usepackage{ragged2e}
\ifCLASSOPTIONcompsoc
  \usepackage[nocompress]{cite}
\else
  \usepackage{cite}
\fi
\ifCLASSINFOpdf
\else
\fi
\usepackage{amssymb}
\newcommand\MYhyperrefoptions{bookmarks=true,bookmarksnumbered=true,
pdfpagemode={UseOutlines},plainpages=false,pdfpagelabels=true,
colorlinks=true,linkcolor={blue},citecolor={blue},urlcolor={blue},
pdftitle={Bare Demo of IEEEtran.cls for Computer Society Journals},%<!CHANGE!
pdfsubject={Typesetting},%<!CHANGE!
pdfauthor={Michael D. Shell},%<!CHANGE!
pdfkeywords={Computer Society, IEEEtran, journal, LaTeX, paper,
             template}}%<^!CHANGE!
\ifCLASSINFOpdf
\usepackage[\MYhyperrefoptions,pdftex]{hyperref}
\else
\usepackage[\MYhyperrefoptions,breaklinks=true,dvips]{hyperref}
\usepackage{breakurl}
\fi

\usepackage{float}

\begin{document}
%
% paper title
% Titles are generally capitalized except for words such as a, an, and, as,
% at, but, by, for, in, nor, of, on, or, the, to and up, which are usually
% not capitalized unless they are the first or last word of the title.
% Linebreaks \\ can be used within to get better formatting as desired.
% Do not put math or special symbols in the title.
\title{Difference-Driven Gating: Adaptive Feature Fusion for U-Net Decoder}
%
%
% author names and IEEE memberships
% note positions of commas and nonbreaking spaces ( ~ ) LaTeX will not break
% a structure at a ~ so this keeps an author's name from being broken across
% two lines.
% use \thanks{} to gain access to the first footnote area
% a separate \thanks must be used for each paragraph as LaTeX2e's \thanks
% was not built to handle multiple paragraphs
%
%
%\IEEEcompsocitemizethanks is a special \thanks that produces the bulleted
% lists the Computer Society journals use for "first footnote" author
% affiliations. Use \IEEEcompsocthanksitem which works much like \item
% for each affiliation group. When not in compsoc mode,
% \IEEEcompsocitemizethanks becomes like \thanks and
% \IEEEcompsocthanksitem becomes a line break with idention. This
% facilitates dual compilation, although admittedly the differences in the
% desired content of \author between the different types of papers makes a
% one-size-fits-all approach a daunting prospect. For instance, compsoc 
% journal papers have the author affiliations above the "Manuscript
% received ..."  text while in non-compsoc journals this is reversed. Sigh.

\author{Kai Li,~\IEEEmembership{Student Member,~IEEE,}
        Xuechao Zou,
        Jiashen Fu,
        Zijun Yan,
        Xintong Wang,
        and~Xiaolin Hu,~\IEEEmembership{Senior Member,~IEEE}% <-this % stops a space
\IEEEcompsocitemizethanks{
\IEEEcompsocthanksitem K. Li and X. Hu are with the Department of Computer Science and Technology, Institute for Artiﬁcial Intelligence, BNRist, IDG/McGovern Institute for Brain Research, Tsinghua University, Beijing, China. X. Hu is also with the Chinese Institute for Brain Research (CIBR), Beijing, China.
\IEEEcompsocthanksitem X. Zou is with the School of Computer Science and Technology, Beijing Jiaotong University, Beijing, China.
\IEEEcompsocthanksitem J. Fu, Z. Yan and X. Wang are with the Department of Computer Science and Technology, Tsinghua University, Beijing, China.
% note need leading \protect in front of \\ to get a newline within \thanks as
% \\ is fragile and will error, could use \hfil\break instead.
% \IEEEcompsocthanksitem J. Fu, Z. Yan and X. Wang contribute equally to the article.
\protect\\
(Corresponding authors: Xiaolin Hu.)

%\IEEEcompsocthanksitem K. Yuan and X. Hu are with the center for Brain-Inspired Computing Research, Tsinghua University, Beijing, China.

}% <-this % stops a space

% \thanks{Manuscript received April 19, 2005; revised August 26, 2015.}
}

% note the % following the last \IEEEmembership and also \thanks - 
% these prevent an unwanted space from occurring between the last author name
% and the end of the author line. i.e., if you had this:
% 
% \author{....lastname \thanks{...} \thanks{...} }
%                     ^------------^------------^----Do not want these spaces!
%
% a space would be appended to the last name and could cause every name on that
% line to be shifted left slightly. This is one of those "LaTeX things". For
% instance, "\textbf{A} \textbf{B}" will typeset as "A B" not "AB". To get
% "AB" then you have to do: "\textbf{A}\textbf{B}"
% \thanks is no different in this regard, so shield the last } of each \thanks
% that ends a line with a % and do not let a space in before the next \thanks.
% Spaces after \IEEEmembership other than the last one are OK (and needed) as
% you are supposed to have spaces between the names. For what it is worth,
% this is a minor point as most people would not even notice if the said evil
% space somehow managed to creep in.

% The paper headers
\markboth{Journal of \LaTeX\ Class Files,~Vol.~14, No.~8, August~2015}%
{Shell \MakeLowercase{\textit{et al.}}: Bare Advanced Demo of IEEEtran.cls for IEEE Computer Society Journals}
% The only time the second header will appear is for the odd numbered pages
% after the title page when using the twoside option.
% 
% *** Note that you probably will NOT want to include the author's ***
% *** name in the headers of peer review papers.                   ***
% You can use \ifCLASSOPTIONpeerreview for conditional compilation here if
% you desire.

% The publisher's ID mark at the bottom of the page is less important with
% Computer Society journal papers as those publications place the marks
% outside of the main text columns and, therefore, unlike regular IEEE
% journals, the available text space is not reduced by their presence.
% If you want to put a publisher's ID mark on the page you can do it like
% this:
%\IEEEpubid{0000--0000/00\$00.00~\copyright~2015 IEEE}
% or like this to get the Computer Society new two part style.
%\IEEEpubid{\makebox[\columnwidth]{\hfill 0000--0000/00/\$00.00~\copyright~2015 IEEE}%
%\hspace{\columnsep}\makebox[\columnwidth]{Published by the IEEE Computer Society\hfill}}
% Remember, if you use this you must call \IEEEpubidadjcol in the second
% column for its text to clear the IEEEpubid mark (Computer Society journal
% papers don't need this extra clearance.)

% use for special paper notices
%\IEEEspecialpapernotice{(Invited Paper)}

% for Computer Society papers, we must declare the abstract and index terms
% PRIOR to the title within the \IEEEtitleabstractindextext IEEEtran
% command as these need to go into the title area created by \maketitle.
% As a general rule, do not put math, special symbols or citations
% in the abstract or keywords.
\IEEEtitleabstractindextext{%
\begin{abstract}
\justifying
The U-Net style models have been widely used in many applications. A critical step in these models is to reconstruct the lower-level features using a top-down decoder. This reconstruction requires precise fusion of high-level semantics and low-level details. Existing attention-based fusion methods typically derive attention weights from the top-down decoder features (global) alone or the correlation between the top-down decoder features and the bottom-up encoder features (local), then modulate the encoder features using these weights. In this work, we explore a different paradigm: deriving attention weights from the difference between the two feature streams. To this end, we propose two difference-based gating approaches: Feature-difference gating (FDG), which directly uses the absolute difference between global and local features to generate adaptive gating maps, and Entropy-difference gating (EDG), which measures the representational certainty of each stream via information entropy and uses their signed entropy difference to derive the attention weights. Both methods produce coupled gating maps that simultaneously modulate the global and local features. Experiments on different tasks including medical image segmentation, remote sensing image cloud removal and speech separation showed that both methods outperformed existing attention-based fusion methods, and EDG performed better. The results suggested a new paradigm for multi-scale feature fusion in the U-Net style structures.
\end{abstract}

% Note that keywords are not normally used for peerreview papers.
}

% make the title area
\maketitle

% To allow for easy dual compilation without having to reenter the
% abstract/keywords data, the \IEEEtitleabstractindextext text will
% not be used in maketitle, but will appear (i.e., to be "transported")
% here as \IEEEdisplaynontitleabstractindextext when compsoc mode
% is not selected <OR> if conference mode is selected - because compsoc
% conference papers position the abstract like regular (non-compsoc)
% papers do!
\IEEEdisplaynontitleabstractindextext
% \IEEEdisplaynontitleabstractindextext has no effect when using
% compsoc under a non-conference mode.

% For peer review papers, you can put extra information on the cover
% page as needed:
% \ifCLASSOPTIONpeerreview
% \begin{center} \bfseries EDICS Category: 3-BBND \end{center}
% \fi
%
% For peerreview papers, this IEEEtran command inserts a page break and
% creates the second title. It will be ignored for other modes.
\IEEEpeerreviewmaketitle

% \ifCLASSOPTIONcompsoc
% \IEEEraisesectionheading{\section{Introduction}\label{sec:introduction}}
% \else
\section{Introduction}
\label{sec:introduction}

\IEEEPARstart{T}{he} U-Net architecture \cite{ronneberger2015u}, a classical encoder-decoder framework, is widely used in various dense prediction tasks, including image processing \cite{oktay2018attnunet,azad2024medical,cai2020maunet,pmaa,shen2014effective} and audio processing \cite{li2022efficient,xu2025tigertimefrequencyinterleavedgain}. This architecture processes information in a bottom-up then top-down manner, analogous to that in biological visual systems~\cite{friston2008hierarchical,pribram2013brain}. It comprises a contracting encoder that abstracts low-level details into high-level semantic representations and an expanding decoder that leverages skip connections to fuse preserved spatial details with the upsampled semantic context~\cite{chen2023auto}.

\begin{figure}[t]
    \centering
    \includegraphics[width=0.9\linewidth]{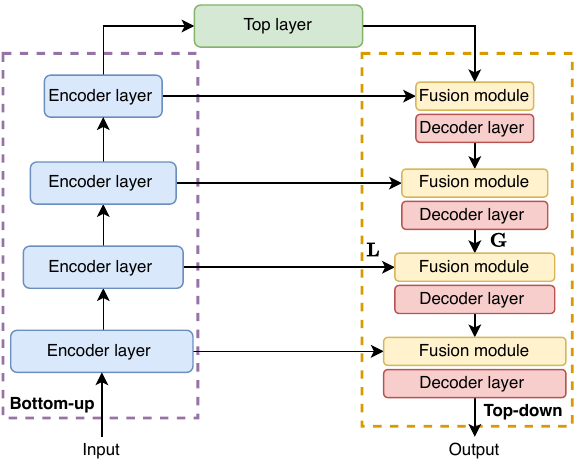}
    \caption{Overview of U-Net architecture with integrated fusion modules. Following the neuroscience convention, the U-Net is depicted in an inverted manner, with the coarsest features placed at the top. Within the fusion modules, $\mathbf{L}$ and $\mathbf{G}$ represent the low-level local features from the encoder and the high-level global features from the decoder, respectively.}
    \label{fig:unet_overall}
\end{figure}

\begin{figure*}[t]
  \centering
  \includegraphics[width=0.85\textwidth]{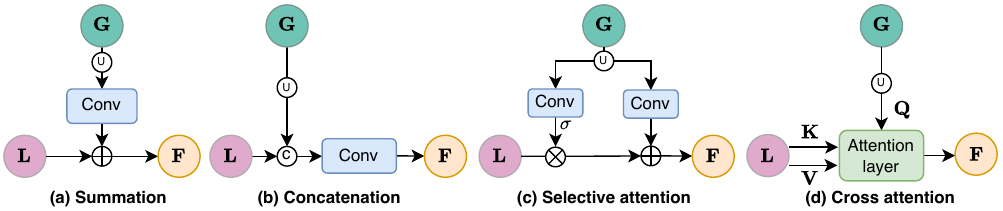}
  \caption{
       Schematic illustration of four representative decoder fusion methods: (a) summation; (b) concatenation; (c) selective attention; (d) cross-attention. Here, $\mathbf{G}$ and $\mathbf{L}$ denote the global features from the top-down decoder and the local features from the bottom-up encoder, respectively. $\mathbf{F}$ represents the fused output feature. The symbols \textcircled{u}, \textcircled{c}, $\oplus$, $\otimes$, and $\sigma$ denote upsampling, concatenation, element-wise summation, element-wise multiplication, and sigmoid activation, respectively.
    }
  \label{fig:fusion_blocks}
\end{figure*}

In the U-Net-based architecture, the effectiveness of the decoder relies on how well it integrates multi-scale information. Therefore, the fusion mechanism that integrates the fine-grained details from the encoder with the semantic context from the decoder becomes a critical bottleneck for performance~\cite{groen2017contributions,skoe2010auditory}. Within the U-Net architecture shown in Fig.~\ref{fig:unet_overall}, this generic fusion mechanism is instantiated at each decoding stage by a fusion module. Existing fusion methods can be broadly divided into two categories.
The first category uses simple fusion mechanisms, which mainly rely on element-wise addition or concatenation~\cite{ronneberger2015u,hu2021speech,tzinis2020sudo}, as shown in Fig.~\ref{fig:fusion_blocks}(a) and (b). Although these methods are computationally efficient, they treat all information indiscriminately and cannot adaptively modulate feature contributions according to contextual relevance. 
The second category employs attention mechanisms, represented by the Attention U-Net~\cite{oktay2018attnunet} and its various extensions~\cite{li2022efficient,xu2025tigertimefrequencyinterleavedgain,pmaa,cai2020maunet,cao2021swinunet}, as shown in Fig.~\ref{fig:fusion_blocks}(c) and (d). Selective attention methods (Fig.~\ref{fig:fusion_blocks}(c)) derive attention weights from the global features and use them to gate the local features $\mathbf{L}$, enabling top-down semantic guidance for feature selection. Cross-attention methods (Fig.~\ref{fig:fusion_blocks}(d)) leverage the correlation between $\mathbf{G}$ and $\mathbf{L}$ through query-key-value projections, capturing richer interactions but incurring substantial computational overhead. While these attention-based methods improve fusion quality, their attention weights are derived from either raw features or pairwise correlations. In this work, we explore whether there exists a more efficient attention mechanism that guides the fusion process.

From the perspective of neuroscience, the encoder-decoder structure resembles feedforward and feedback pathways in the visual and auditory systems  \cite{hickok2007corticalspeech}, \cite{bastos2012circuitsforpc}. A computational theory, predictive coding \cite{rao1999pc}, claims that a core function of feedback connections between functional areas in the cortex is to predict the lower-level representations and compute the prediction error. This error signal serves different purposes in computational models \cite{rao1999pc}, \cite{spratling2010pc}, \cite{hanDeepPredictiveCoding2018}. This inspires us to build attention for fusing the global and local features based on the error signal in the U-Net, considering that the decoder is essentially making predictions for lower-level representations.

Towards this goal, we propose two difference-based gating approaches. \emph{Feature-difference gating (FDG)} module computes the absolute difference between encoder and decoder features and converts it into adaptive gating maps that simultaneously modulate both $\mathbf{G}$ and $\mathbf{L}$. Despite its simplicity, our experiments showed that the FDG module already outperformed existing attention-based fusion methods (see Sec.~\ref{sec:comp_fusion}).
\emph{Entropy-Difference Gating} (EDG) takes a different approach by quantifying the representational certainty of each feature stream via information entropy~\cite{shannon1948mathematical}. Specifically, the signed entropy difference between the global and local streams serves as the gating signal along both channel and spatiotemporal dimensions, adaptively biasing the fusion toward the lower-entropy (more certain) stream. Like the FDG module, the EDG module produces coupled gating maps for both streams, enabling simultaneous modulation of $\mathbf{G}$ and $\mathbf{L}$ with negligible additional cost.

We integrated both the FDG and EDG modules into existing U-Net-based architectures. Experimental results on different tasks including the medical image segmentation, remote sensing image cloud removal and speech separation showed that the simple FDG module enabled the U-Net models to achieve much better results than previous state-of-the-art models, and that the EDG module-based models achieved even better results. 

% \fi
% \IEEEPARstart{H}{uman} 

\section{Preliminaries and Related Work}
\label{sec:related_works}

The U-Net architecture \cite{ronneberger2015u} and its variants \cite{zhou2018unet++,oktay2018attnunet,azad2024medical,li2022efficient} (Fig.~\ref{fig:unet_overall}) are widely used in dense prediction tasks across medical imaging, remote sensing, and speech separation. A critical component of these architectures is the feature fusion module within the decoder, which bridges the gap between global features from the top-down decoder and local features from the bottom-up encoder.

\subsection{Simple Fusion Methods}
Summation (Fig. 2(a)) and concatenation (Fig. 2(b)) are two simple fusion methods. In Figs.~\ref{fig:fusion_blocks}(a) and (b), we use $\mathbf{G}$ and $\mathbf{L}$ to denote the global features from the top-down decoder and the local features from the bottom-up encoder, respectively. The summation method aligns the channel dimensions of the upsampled global features and local encoder features (typically via $1\times 1$ convolution) and fuses them via element-wise addition. While computationally efficient, this method, employed in variants like Res-UNet~\cite{diakogiannis2020resunet}, assumes that features from different semantic levels contribute equally to the reconstruction. Alternatively, the concatenation method (Fig.~\ref{fig:fusion_blocks}(b)), adopted by the original U-Net \cite{ronneberger2015u}, stacks features along the channel dimension, followed by convolutional layers for feature mixing. Although it preserves more information than summation, both methods treat the incoming signal streams indiscriminately. They lack the adaptivity to suppress noise or emphasize task-relevant regions, which is particularly detrimental when the local encoder features contain significant background noise or artifacts \cite{hu2021speech,tzinis2020sudo}.

\subsection{Attention-Based Fusion Methods}
To overcome the limitations of simple fusion, recent methods have introduced attention mechanisms to actively modulate feature fusion, as illustrated in Fig.~\ref{fig:fusion_blocks}(c) and (d).
Selective attention (Fig.~\ref{fig:fusion_blocks}(c)), widely adopted in models such as AttnUNet~\cite{oktay2018attnunet}, PMAA~\cite{pmaa}, and TDANet~\cite{li2022efficient}, derives attention weights solely from the global features $\mathbf{G}$ (e.g., via convolution followed by a sigmoid activation) and uses them to multiplicatively modulate the local features $\mathbf{L}$. This top-down gating suppresses irrelevant regions before fusion, but only $\mathbf{L}$ is modulated while $\mathbf{G}$ serves solely as a conditioning signal without being adjusted itself.
Cross-attention (Fig.~\ref{fig:fusion_blocks}(d)), inspired by the success of Transformers~\cite{fan2020manet,cao2021swinunet}, takes the upsampled global feature as the query and the local feature as both key and value. This formulation captures richer interactions by leveraging the correlation between $\mathbf{G}$ and $\mathbf{L}$, but incurs substantial computational overhead. Moreover, similar to selective attention, the output is derived primarily from $\mathbf{L}$. In both cases, the attention weights are computed from either raw features or pairwise correlations, without explicitly reflecting the differences between the two streams.

\section{Difference-Based Gating Module}
\label{sec:methods}

\subsection{Overall Pipeline}

\begin{figure}[t]
    \centering
    \includegraphics[width=0.75\linewidth]{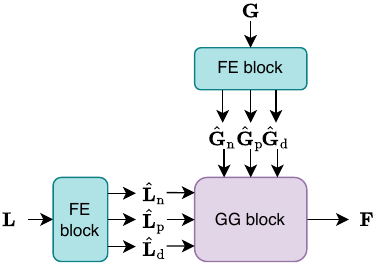}
    \caption{Overall pipeline of the proposed gating modules. The two FE blocks share the same structure with different parameters. The FDG and EDG modules differ only in how the FE block compresses intermediate features: FDG uses mean pooling to produce feature statistics, while EDG computes Shannon entropy to quantify representational certainty. The subsequent GG block processing is identical for both modules.}
    \label{fig:edg_overall}
\end{figure}

As shown in Fig.~\ref{fig:edg_overall}, the proposed gating modules replace the fusion module in the U-Net decoder. Unlike existing fusion methods, they dynamically modulate the contributions of both the global and local features by quantifying the \emph{difference} between the two streams. We present two instantiations, FDG and EDG, that share the same overall pipeline and differ only in how the FE block compresses intermediate features (Secs.~\ref{sec:fdg} and~\ref{sec:edg}). The subsequent gating generation procedure (Sec.~\ref{sec:gg_module}) is identical for both modules.

The module consists of two main components: a feature extraction (FE) block and a gating generation (GG) block. Let $\mathbf{G}\in \mathbb{R}^{C\times P}$ denote the global features from the top-down decoder, and let $\mathbf{L}\in \mathbb{R}^{C\times P'}$ represent the local features from the bottom-up encoder. For notational simplicity, we assume that both $\mathbf{G}$ and $\mathbf{L}$ have $C$ channels. The module employs a unified framework applicable to both visual and auditory tasks, where the only difference lies in the interpretation of the second dimension. In computer vision tasks, $P$ and $P'$ correspond to the number of spatial locations (pixels), typically with $P < P'$, whereas in audio processing tasks they usually denote the number of time steps or time–frequency bins.

Specifically, for the global decoder features $\mathbf{G}\in\mathbb{R}^{C\times P}$, the FE block produces $\hat{\mathbf{G}}_{\text{n}}\in\mathbb{R}^{C\times 1}$, $\hat{\mathbf{G}}_{\text{p}}\in\mathbb{R}^{1\times P}$, and $\hat{\mathbf{G}}_{\text{d}}\in\mathbb{R}^{C\times P}$. Similarly, for the local encoder features $\mathbf{L}\in\mathbb{R}^{C\times P'}$, it outputs $\hat{\mathbf{L}}_{\text{n}}\in\mathbb{R}^{C\times 1}$, $\hat{\mathbf{L}}_{\text{p}}\in\mathbb{R}^{1\times P'}$, and $\hat{\mathbf{L}}_{\text{d}}\in\mathbb{R}^{C\times P'}$. The GG block then takes $\{\hat{\mathbf{G}}_{\text{n}}, \hat{\mathbf{G}}_{\text{p}}, \hat{\mathbf{G}}_{\text{d}}\}$ and $\{\hat{\mathbf{L}}_{\text{n}}, \hat{\mathbf{L}}_{\text{p}}, \hat{\mathbf{L}}_{\text{d}}\}$ and computes difference metrics along both the channel and spatiotemporal dimensions to derive adaptive gating maps. Finally, the upsampled global content feature $\hat{\mathbf{G}}_{\text{d}}$ and the local content feature $\hat{\mathbf{L}}_{\text{d}}$ are modulated by the resulting gates and fused to yield $\mathbf{F}\in\mathbb{R}^{C\times P'}$.

\subsection{Feature Extraction Block}
\begin{figure}[t]
    \centering
    \includegraphics[width=0.9\linewidth]{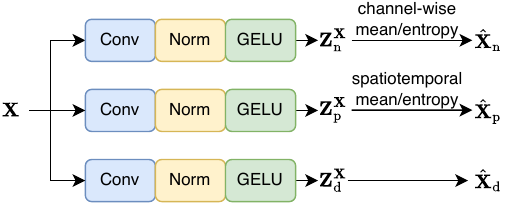}
    \caption{Overall pipeline of the FE block.}
    \label{fig:ee_all}
\end{figure}

The FE block operates on a single feature stream at a time. It produces two complementary compressed features for the subsequent GG block, along with a refined content feature, as shown in Fig.~\ref{fig:ee_all}. The block processes an input feature map $\mathbf{X} \in \mathbb{R}^{C \times Q}$ with $\mathbf{X}\in\{\mathbf{G},\mathbf{L}\}$, where $Q=P$ for $\mathbf{G}$ and $Q=P'$ for $\mathbf{L}$.
Formally, the FE block processing is expressed as:
\begin{equation}
\big\{\hat{\mathbf{X}}_{\text{n}}, \hat{\mathbf{X}}_{\text{p}}, \hat{\mathbf{X}}_{\text{d}}\big\} = \mathrm{FE}(\mathbf{X};\theta),
\end{equation}
where the block yields three outputs: $\hat{\mathbf{X}}_{\text{n}}\in\mathbb{R}^{C\times 1}$ (a per-channel feature obtained by compressing along the spatiotemporal dimension), $\hat{\mathbf{X}}_{\text{p}}\in\mathbb{R}^{1\times Q}$ (a per-position feature obtained by compressing along the channel dimension), and $\hat{\mathbf{X}}_{\text{d}}\in\mathbb{R}^{C\times Q}$ (refined content feature for fusion). The specific form of $\hat{\mathbf{X}}_{\text{n}}$ and $\hat{\mathbf{X}}_{\text{p}}$ depends on the aggregation function used in the gating mode: in the FDG module, mean pooling serves as the aggregation function, producing aggregated feature statistics (Sec.~\ref{sec:fdg}), while in the EDG module, Shannon entropy is used instead, yielding entropy scores (Sec.~\ref{sec:edg}).

Specifically, the FE block adopts a three-branch architecture. First, we apply three independent $1\times 1$ convolutional layers, each followed by Batch Normalization~\cite{ioffe2015batch} and GELU activation~\cite{hendrycks2016gaussian}, to generate intermediate feature maps
$\mathbf{Z}_\text{n}^{\mathbf{X}}\in\mathbb{R}^{C\times Q}$,
$\mathbf{Z}_\text{p}^{\mathbf{X}}\in\mathbb{R}^{C\times Q}$, and
$\mathbf{Z}_\text{d}^{\mathbf{X}}\in\mathbb{R}^{C\times Q}$ for the three branches, respectively. The first two branches ($\mathbf{Z}_\text{n}^{\mathbf{X}}$ and $\mathbf{Z}_\text{p}^{\mathbf{X}}$) are then aggregated into $\hat{\mathbf{X}}_{\text{n}}$ and $\hat{\mathbf{X}}_{\text{p}}$, respectively, following mode-specific procedures described in Secs.~\ref{sec:fdg} and~\ref{sec:edg}.
For the third branch, the intermediate feature map directly serves as the refined content feature, i.e., $\hat{\mathbf{X}}_{\text{d}} = \mathbf{Z}_\text{d}^{\mathbf{X}}$.

\subsubsection{Mean-Pooling Aggregation (FDG)}
\label{sec:fdg}

In the FDG module, the first two branches aggregate the intermediate features via mean pooling along complementary dimensions to obtain compact features:
\begin{align}
\hat{\mathbf{X}}_{\text{n}}
    &= \text{Mean}_{q}\big(\mathbf{Z}_{\text{n}}^{\mathbf{X}}\big)
       \in\mathbb{R}^{C\times 1}, \label{eq:fdg_n}\\
\hat{\mathbf{X}}_{\text{p}}
    &= \text{Mean}_{c}\big(\mathbf{Z}_{\text{p}}^{\mathbf{X}}\big)
       \in\mathbb{R}^{1\times Q}. \label{eq:fdg_p}
\end{align}
Here, $\hat{\mathbf{X}}_{\text{n}}$ summarizes the average activation of each channel across all spatiotemporal positions, while $\hat{\mathbf{X}}_{\text{p}}$ captures the average response at each position across all channels. These compact features serve as the basis for measuring feature-level disagreement between the two streams in the GG block.

\subsubsection{Entropy-Based Aggregation (EDG)}
\label{sec:edg}

In the EDG module, the first two branches replace mean pooling with Shannon entropy~\cite{shannon1948mathematical} computed along complementary dimensions, thereby capturing the representational certainty of each feature stream. Let $\varepsilon>0$ be a constant for numerical stability.

For the spatiotemporal entropy, we treat, for each channel $c$, the vector $\mathbf{Z}_{\text{n}}^{\mathbf{X}}(c,:)$ as a distribution over the spatiotemporal dimension $Q$ and apply softmax along $Q$ to obtain $\mathbf{P}_\text{n}^{\mathbf{X}}\in\mathbb{R}^{C\times Q}$:
\begin{equation}
\mathbf{P}_{\text{n}}^{\mathbf{X}}(c,q)
= \frac{\exp\big(\mathbf{Z}_{\text{n}}^{\mathbf{X}}(c,q)\big)}
        {\sum_{q'=1}^{Q}\exp\big(\mathbf{Z}_{\text{n}}^{\mathbf{X}}(c,q')\big)+\varepsilon}.
\end{equation}
The spatiotemporal entropy is then computed by summing over $Q$ for each channel:
\begin{equation}
\hat{\mathbf{X}}_{\text{n}}(c)
= -\sum_{q=1}^{Q} \mathbf{P}_{\text{n}}^{\mathbf{X}}(c,q)
     \log\big(\mathbf{P}_{\text{n}}^{\mathbf{X}}(c,q)\big),
\label{eq:en}
\end{equation}
yielding $\hat{\mathbf{X}}_{\text{n}}\in\mathbb{R}^{C\times 1}$.
For the channel-wise entropy, we interpret, for each location $q$, the vector $\mathbf{Z}_{\text{p}}^{\mathbf{X}}(:,q)$ as a distribution over channels and apply softmax along the channel dimension to obtain $\mathbf{P}_\text{p}^{\mathbf{X}}\in\mathbb{R}^{C\times Q}$:
\begin{equation}
\mathbf{P}_{\text{p}}^{\mathbf{X}}(c,q)
= \frac{\exp\big(\mathbf{Z}_{\text{p}}^{\mathbf{X}}(c,q)\big)}
        {\sum_{c'=1}^{C}\exp\big(\mathbf{Z}_{\text{p}}^{\mathbf{X}}(c',q)\big)+\varepsilon}.
\end{equation}
The channel-wise entropy is then computed by summing over channels at each position:
\begin{equation}
\hat{\mathbf{X}}_{\text{p}}(q)
= -\sum_{c=1}^{C} \mathbf{P}_{\text{p}}^{\mathbf{X}}(c,q)
     \log\big(\mathbf{P}_{\text{p}}^{\mathbf{X}}(c,q)\big),
\label{eq:ep}
\end{equation}
which gives $\hat{\mathbf{X}}_{\text{p}}\in\mathbb{R}^{1\times Q}$. Lower entropy indicates that the feature activations are concentrated on a few positions (or channels), reflecting higher representational certainty; higher entropy implies a more diffuse, less decisive distribution.

\subsubsection{FE Block Instantiation}
\label{sec:fe_inst}

As shown in Fig.~\ref{fig:edg_overall}, the FE block is applied to both feature streams independently:
\begin{align}
\{\hat{\mathbf{G}}_{\text{n}}, \hat{\mathbf{G}}_{\text{p}}, \hat{\mathbf{G}}_{\text{d}}\}
    &= \mathrm{FE}(\mathbf{G};\theta^{G}), \label{eq:ee_g} \\
\{\hat{\mathbf{L}}_{\text{n}}, \hat{\mathbf{L}}_{\text{p}}, \hat{\mathbf{L}}_{\text{d}}\}
    &= \mathrm{FE}(\mathbf{L};\theta^{L}). \label{eq:ee_l}
\end{align}
The resulting six tensors are then passed to the GG block for difference-based gating.

\subsection{Gating Generation Block}
\label{sec:gg_module}

\begin{figure}[t]
    \centering
    \includegraphics[width=0.85\linewidth]{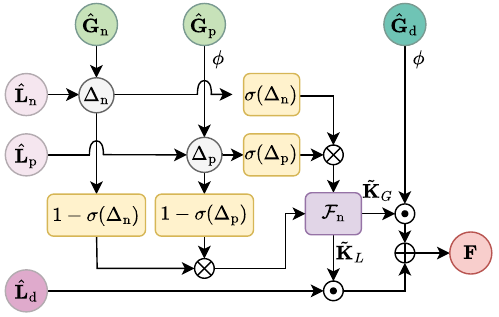}
    \caption{Architecture of the GG block. In the FDG module, $\Delta_{\text{n}}$ and $\Delta_{\text{p}}$ are instantiated as absolute feature differences; in the EDG module, they are defined as signed entropy differences that reflect relative certainty.}
    \label{fig:gg_module}
\end{figure}

The GG block is the decision-making component that determines fusion weights by evaluating the difference between the global and local compressed feature pairs produced by the FE blocks. As shown in Fig.~\ref{fig:gg_module}, it follows a unified processing framework for both visual and auditory tasks. The block receives the six feature tensors from the preceding FE blocks: the global-stream components $\{\hat{\mathbf{G}}_{\text{n}}, \hat{\mathbf{G}}_{\text{p}}, \hat{\mathbf{G}}_{\text{d}}\}$ and the local-stream components $\{\hat{\mathbf{L}}_{\text{n}}, \hat{\mathbf{L}}_{\text{p}}, \hat{\mathbf{L}}_{\text{d}}\}$. Since $P < P'$ (the global features have lower resolution), resolution alignment via nearest-neighbor interpolation $\phi(\cdot)$ is applied before element-wise comparison. The GG block first computes variant-specific difference metrics $\Delta_{\text{n}}$ and $\Delta_{\text{p}}$, then transforms them into gating maps through a shared pipeline.

\subsubsection{Difference metrics}
The FDG and EDG modules differ only in how the difference metrics are computed from the compressed features.

In the FDG module, the metrics measure the absolute feature-level disagreement between the two streams:
\begin{align}
\Delta_{\text{n}} &= \big\lvert \hat{\mathbf{G}}_{\text{n}} - \hat{\mathbf{L}}_{\text{n}} \big\rvert \in \mathbb{R}^{C\times 1}, \label{eq:deltan_feat} \\
\Delta_{\text{p}} &= \big\lvert \phi(\hat{\mathbf{G}}_{\text{p}}) - \hat{\mathbf{L}}_{\text{p}} \big\rvert \in \mathbb{R}^{1\times P'}. \label{eq:deltap_feat}
\end{align}
The absolute value is used because two values with the same magnitude but different signs should represent the same level of feature difference. $\Delta_{\text{n}}$ provides, for each channel, a compressed difference feature aggregated over space or time, while $\Delta_{\text{p}}$ summarizes the difference strength at each position across channels. Since these quantities are non-negative by construction, they naturally encode the magnitude of the disagreement.

In the EDG module, the metrics instead capture the signed certainty difference based on the entropy scores:
\begin{align}
\Delta_{\text{n}} &= \hat{\mathbf{G}}_{\text{n}} - \hat{\mathbf{L}}_{\text{n}}, \label{eq:dn_ent} \\
\Delta_{\text{p}} &= \phi(\hat{\mathbf{G}}_{\text{p}}) - \hat{\mathbf{L}}_{\text{p}}. \label{eq:dp_ent}
\end{align}
These signed differences intuitively reflect the relative reliability of the two streams:
\begin{itemize}
\item \textbf{Negative values} ($\Delta<0$) indicate that the global features have lower entropy and are thus more certain, so they should be prioritized.
\item \textbf{Positive values} ($\Delta>0$) indicate that the local features have lower entropy, suggesting that the local features are more reliable and should receive higher fusion priority.
\end{itemize}

\subsubsection{Shared gating pipeline}
Given $\Delta_{\text{n}}$ and $\Delta_{\text{p}}$ (instantiated as either the feature-difference or entropy-difference metrics above), the GG block generates two multiplicative gating maps for the local and global streams:
\begin{align}
\mathbf{K}_{L} &= \sigma(\Delta_{\text{n}}) \otimes \sigma(\Delta_{\text{p}}) \in \mathbb{R}^{C \times P'}, \label{eq:kl_general} \\
\mathbf{K}_{G} &= \big(1 - \sigma(\Delta_{\text{n}})\big) \otimes \big(1 - \sigma(\Delta_{\text{p}})\big) \in \mathbb{R}^{C \times P'}, \label{eq:kg_general}
\end{align}
where $\sigma(\cdot)$ denotes the sigmoid function and $\otimes$ here denotes broadcast multiplication along complementary dimensions. In the FDG module, $\Delta$ is non-negative, so $\sigma(\Delta)\geq 0.5$: large disagreement biases the fusion toward the local stream, while small disagreement yields approximately equal weighting. This embeds an inductive bias that encoder features that preserve richer spatial detail should be preferred when the two streams diverge. In the EDG module, $\Delta$ is signed, allowing $\sigma(\Delta)$ to span $(0,1)$ and adaptively favor whichever stream has lower entropy, making the gating direction data-driven rather than relying on a fixed prior.

The gating maps are then normalized by $\mathcal{F}_{\text{n}}$ (see Fig.~\ref{fig:gg_module}):
\begin{equation}
(\tilde{\mathbf{K}}_{G},\; \tilde{\mathbf{K}}_{L}) = \mathcal{F}_{\text{n}}(\mathbf{K}_{G},\; \mathbf{K}_{L}) = \left(\frac{\mathbf{K}_{G}}{\mathbf{S}},\; \frac{\mathbf{K}_{L}}{\mathbf{S}}\right),
\label{eq:gate_norm}
\end{equation}
where $\mathbf{S} = \mathbf{K}_{G} + \mathbf{K}_{L} + \varepsilon$, the division is element-wise, and $\varepsilon$ is a small constant for numerical stability. Since both maps share the same denominator $\mathbf{S}$, the ordering between $\tilde{\mathbf{K}}_{G}$ and $\tilde{\mathbf{K}}_{L}$ at each position is solely determined by $\Delta_{\text{n}}$ and $\Delta_{\text{p}}$, enabling joint channel-spatiotemporal gating.

The normalized gating maps $\tilde{\mathbf{K}}_{G}$ and $\tilde{\mathbf{K}}_{L}$ are applied to the content features $\hat{\mathbf{G}}_{\text{d}}$ and $\hat{\mathbf{L}}_{\text{d}}$ output by the FE blocks. The final fused feature $\mathbf{F} \in \mathbb{R}^{C \times P'}$ is computed as
\begin{equation}
\mathbf{F} = \phi(\hat{\mathbf{G}}_{\text{d}}) \odot \tilde{\mathbf{K}}_{G}
           + \hat{\mathbf{L}}_{\text{d}} \odot \tilde{\mathbf{K}}_{L},
\end{equation}
where $\odot$ denotes element-wise multiplication. Unlike existing attention-based fusion methods that only modulate one feature stream (as shown in Fig.~\ref{fig:fusion_blocks}(c) and (d)), the FDG and EDG modules modulate both the global and local content features through their respective gating maps, enabling a more balanced fusion. Since the two modules differ only in how $\Delta_{\text{n}}$ and $\Delta_{\text{p}}$ are computed while the content features being gated and the subsequent fusion procedure are identical, this design ensures a fair comparison in the ablation study (Sec.~\ref{sec:comp_fusion}).

\begin{figure}[t]
    \centering
    \includegraphics[width=0.8\linewidth]{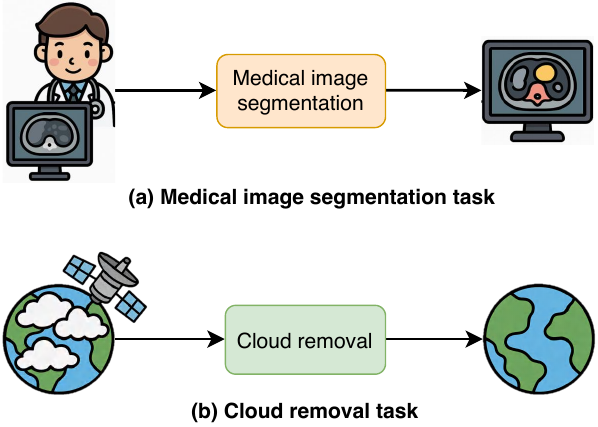}
    \caption{Overall processing pipelines for the medical image segmentation and cloud-removal tasks.}
    \label{fig:visual_pipeline}
\end{figure}
\begin{figure*}[t]
\centering
\includegraphics[width=1.0\textwidth]{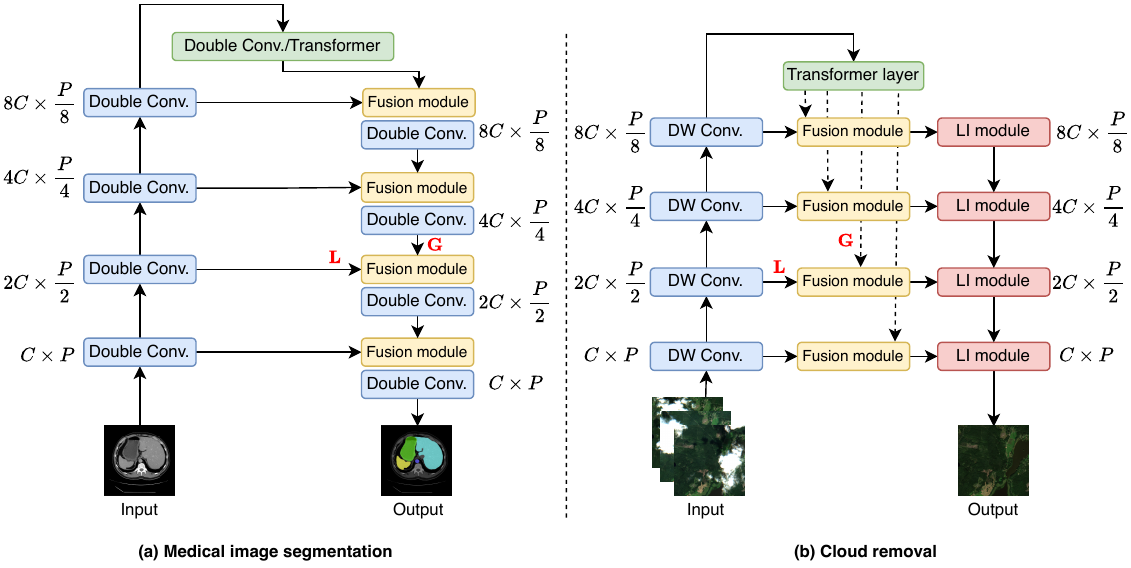}
\caption{Network architectures for two vision tasks. (a) Medical image segmentation based on U-Net~\cite{ronneberger2015u} and TransAttUNet~\cite{chen2022transattunet} architectures, where the original fusion modules are replaced by EDG modules. The major difference between these two architectures lies in their global feature processing at the top level: U-Net uses convolutional layers, whereas TransAttUNet employs Transformer layers. (b) Cloud removal based on the U-Net-like PMAA model~\cite{pmaa}, where the fusion modules are replaced by EDG modules. In (b), dashed arrows denote the shared top-level feature that is fed as the global input to each EDG module; each EDG instance thus receives one global feature stream and one local feature stream. The dimensions annotated in the diagrams represent the output dimensions of the convolutional layers.}
\label{fig:med_pcunet}
\end{figure*}

\subsection{Domain-Specific Integration}
The proposed gating modules are designed as general-purpose fusion components for U-Net-based architectures. Below we describe its integration into three representative tasks across image and audio domains: medical image segmentation, remote sensing image cloud removal, and speech separation. In this section, we mainly introduce the integration of the EDG module, and the integration of the FDG module is similar.

\subsubsection{Applications to Image Processing}

For medical image segmentation (Fig.~\ref{fig:visual_pipeline}(a)), we adopt the U-Net~\cite{ronneberger2015u} and TransAttUNet~\cite{chen2022transattunet} as baseline models. As shown in Fig.~\ref{fig:med_pcunet}(a), both architectures share a similar symmetric encoder–decoder backbone and employ skip connections between corresponding encoder and decoder layers. The encoder consists of cascaded stages, each applying convolutional layers with normalization and activation, then downsampling to reduce spatial resolution and capture broader context. Correspondingly, the decoder upsamples features stage by stage to gradually recover spatial resolution. The primary distinction between these two models lies in their bottleneck design: U-Net relies on standard convolutions for high-level abstraction, while TransAttUNet introduces Transformer layers to capture long-range dependencies. However, both models employ simple channel-wise concatenation to fuse features from skip connections and corresponding decoder layers, treating the global and local features equally without considering their relative reliability~\cite{zhang2018exfuse}.

We replace the standard concatenation fusion modules in both models with our EDG module (Fig.~\ref{fig:med_pcunet}(a)). 
Let $\mathbf{G}\in \mathbb{R}^{C\times P}$ denote the global features from the decoder, and $\mathbf{L}\in \mathbb{R}^{C\times P'}$ denote the corresponding local features from the encoder. The EDG module takes $(\mathbf{G}, \mathbf{L})$ as input, producing a fused representation. Apart from substituting the fusion block with the EDG module, we keep the network depth, per-layer channel configuration, and supervision heads identical to the original baseline models.

For remote sensing image cloud removal (Fig.~\ref{fig:visual_pipeline}(b)), the goal is to reconstruct a clean image from multiple cloud-contaminated inputs captured at different times. We adopt PMAA~\cite{pmaa} as the baseline (Fig.~\ref{fig:med_pcunet}(b)), which employs a U-Net-style encoder--decoder with Transformer layers for global context modeling. Its decoder integrates a local interaction (LI) module that uses selective attention to fuse features. Let $\mathbf{G}$ denote the global features from the Transformer and $\mathbf{L}$ denote the local features from the lateral connection. The selective-attention fusion in PMAA is formulated as
\begin{equation}
\mathbf{L}' = \phi\big(\sigma\big(\mathbf{Z}_1(\mathbf{G})\big)\big)\odot\mathbf{Z}_2(\mathbf{L}) + \phi\big(\mathbf{Z}_3(\mathbf{G})\big),
\end{equation}
where $\mathbf{Z}_1(\cdot), \mathbf{Z}_2(\cdot)$, and $\mathbf{Z}_3(\cdot)$ are linear layers. This mechanism modulates $\mathbf{L}$ using $\mathbf{G}$ as a top-down gating signal, but the interaction is one-way: $\mathbf{L}$ cannot inform the fusion about its own reliability. We replace this selective-attention block with the EDG module (Sec.~\ref{sec:edg}).

\begin{figure}[t]
    \centering
    \includegraphics[width=0.8\linewidth]{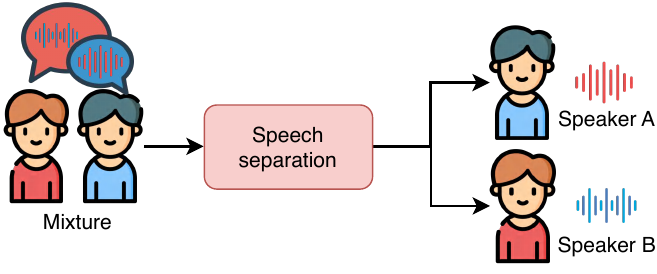}
    \caption{Overall processing pipeline for the speech separation task.}
    \label{fig:ss_pipeline}
\end{figure}

\begin{figure}[t]
    \centering
    \includegraphics[width=1.0\linewidth]{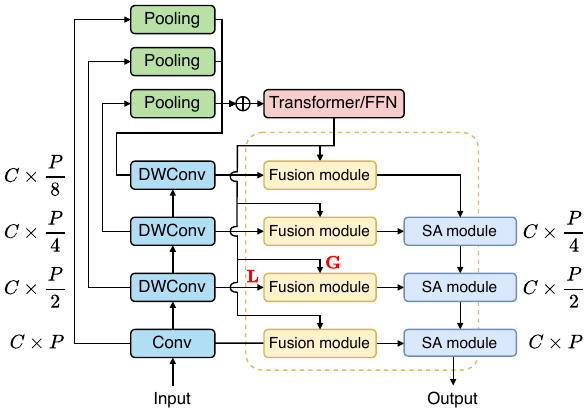}
    \caption{The diagram illustrates the separator structure of TDANet \cite{li2022efficient} and the Multi-Scale Attention (MSA) module within TIGER's separator \cite{xu2025tigertimefrequencyinterleavedgain}. While these specific sub-modules share a similar U-Net-like backbone, their top-level global processing differs (a Transformer for TDANet and a feed-forward network for TIGER). Their original fusion modules, which rely on a selective attention mechanism, are replaced with the proposed EDG module. The dimensions annotated in the diagrams represent the output sizes of the convolutional layers.}
    \label{fig:audio_arch}
\end{figure}

\subsubsection{Applications to Audio Processing}

In this work, we consider a specific audio processing application, speech separation. The objective of speech separation is to accurately recover the speech of individual speakers from a mixed signal, as illustrated in Fig.~\ref{fig:ss_pipeline}. Existing lightweight speech separation models typically employ U-Net-based architectures, such as TDANet~\cite{li2022efficient} and TIGER~\cite{xu2025tigertimefrequencyinterleavedgain}. An encoder first extracts multi-scale features, which are then used by a decoder to reconstruct the separated speech signals for different speakers. These models fuse global and local features during the decoder’s top-down reconstruction, often utilizing a selective attention (SA) mechanism, as shown in the fusion block of Fig.~\ref{fig:audio_arch}. Specifically, let the global features from the Transformer or FFN be denoted as $\mathbf{G}$, and the local features from the skip connection be denoted as $\mathbf{L}$. The SA mechanism projects the global features $\mathbf{G}$ into a gating feature $\mathbf{G}_{\text{gate}}$ and a residual feature $\mathbf{G}_{\text{res}}$. Since the global features have lower resolution, they are upsampled to match the dimensions of the local features $\mathbf{L}$. The final fusion process is expressed as
\begin{equation}
\mathbf{L}_{\text{fused}} = \sigma(\phi(\mathbf{G}_{\text{gate}})) \odot \mathbf{L} + \phi(\mathbf{G}_{\text{res}}).
\label{eq:sa_fusion}
\end{equation}

This SA mechanism applies one-way gating from the decoder to the encoder stream. We replace it with the EDG module (Sec.~\ref{sec:edg}).

\section{Experiments}
\label{sec:exps}

We evaluated the proposed method on three tasks spanning distinct data modalities: medical image segmentation (Sec.~\ref{sec:med_seg}), cloud removal (Sec.~\ref{sec:cloud_removal}), and speech separation (Sec.~\ref{sec:speech_sep}). Both FDG and EDG were evaluated on all three tasks; for brevity, we report FDG results only on medical image segmentation (Sec.~\ref{sec:comp_fusion}), where a controlled comparison showed that both modules outperformed existing attention-based fusion methods and EDG performed better. The EDG module was therefore adopted as the default configuration for all subsequent experiments. Ablation studies were presented in Sec.~\ref{sec:ablation}. Unless otherwise specified, baseline results were taken from the original publications or reproduced under identical training settings when official results were unavailable. The source codes will be made publicly available upon acceptance of the paper.

\subsection{Medical Image Segmentation}
\label{sec:med_seg}

\begin{table}[t]
    \centering
    \caption{Performance comparison of different fusion methods on the Synapse dataset using U-Net~\cite{ronneberger2015u}. The best results are in bold. All results are reported as mean $\pm$ standard deviation over 10 independent runs.}
    \label{tab:fusion_comparison}
    \renewcommand{\arraystretch}{1.2}
    \resizebox{0.4\textwidth}{!}{
    \begin{tabular}{l|cc}
        \toprule
        \textbf{Fusion Methods} & \textbf{Avg. DSC}~$\uparrow$ & \textbf{HD95}~$\downarrow$ \\
        \midrule
        Summation                        & 79.54$\pm$0.34 & 20.08$\pm$1.31 \\
        Concatenation                    & 79.98$\pm$0.32 & 25.91$\pm$1.47 \\
        Cross-attention                  & 76.60$\pm$0.41 & 32.74$\pm$1.63 \\
        Selective-attention              & 79.56$\pm$0.33 & 28.36$\pm$1.38 \\
        \midrule
        \rowcolor[RGB]{217,217,217} FDG module (Ours)    & 81.45$\pm$0.31 & 16.93$\pm$1.08 \\
        \rowcolor[RGB]{217,217,217} \textbf{EDG module (Ours)} & \textbf{82.96$\pm$0.27} & \textbf{14.52$\pm$0.96} \\
        \bottomrule
    \end{tabular}
    }
\end{table}

\begin{table}[t]
    \centering
    \caption{Comparison of computational efficiency of different fusion methods in U-Net~\cite{ronneberger2015u}.}
    \label{tab:fusion_comparison_unet_pmaa}
    \renewcommand{\arraystretch}{1.2}
    \resizebox{0.5\textwidth}{!}{
    \begin{tabular}{l|cccc}
        \toprule
        \textbf{Fusion Methods} & \textbf{Params (M)} & \textbf{MACs (G)} & \textbf{CPU (ms)} & \textbf{GPU (ms)} \\
        \midrule
        Summation                 & 30.36 & 40.45 & 54.31 & 3.00 \\
        Concatenation              & 32.10 & 44.59 & 54.63 & 3.22 \\
        Cross-attention     & 32.46 & 80.40 & 185.34 & 16.41 \\
        Selective-attention & 33.76 & 42.79 & 53.06 & 3.34 \\
        \rowcolor[RGB]{217,217,217} FDG module (Ours) & 34.56 & 42.96 & 55.18 & 3.54 \\
        \rowcolor[RGB]{217,217,217} \textbf{EDG module (Ours)} & 34.56 & 43.09 & 55.47 & 3.59 \\
        \bottomrule
    \end{tabular}
    }
\end{table}

\begin{table*}[t]
  \centering
  \caption{Performance comparison on the Synapse dataset. Metrics include mean Dice coefficient (\%) and HD95, alongside per-class Dice scores. All models were trained under identical settings for fair comparison. \textbf{Bold} indicates the best result, while \underline{underlined} denotes the second best. The numbers in red indicate absolute improvements (in pp or mm) relative to the original TransAttUNet and U-Net baselines.}
  \label{tab:med_synapse}
  \renewcommand{\arraystretch}{1.2}
  \resizebox{1.0\textwidth}{!}{
  \begin{tabular}{l|cc|cccccccc}
  \toprule
  \textbf{Model} & \textbf{Avg. DSC $\uparrow$} & \textbf{HD95 $\downarrow$} & \textbf{Aorta $\uparrow$} & \textbf{Gallbladder $\uparrow$} & \textbf{Left kidney $\uparrow$} & \textbf{Right kidney $\uparrow$} & \textbf{Liver $\uparrow$} & \textbf{Pancreas $\uparrow$} & \textbf{Spleen $\uparrow$} & \textbf{Stomach $\uparrow$} \\ 
  \midrule
  UNext\cite{valanarasu2022unext}                & 68.85             & 30.82             & 80.01          & 46.65          & 75.69          & 69.38          & 92.35          & 38.77          & 83.42          & 64.56 \\
 UNet++\cite{zhou2018unetplusplus}& 81.55& 18.81& 89.46& 66.12& 84.92& \underline{82.25}& 94.94& 65.93& 89.86&78.90\\
  AttnUNet \cite{oktay2018attnunet}              & 81.92& \underline{16.76}& 88.58& \textbf{68.83}& 85.22& 78.20& 94.48& 67.90& \textbf{91.94}& 80.25\\
  MANet\cite{fan2020manet}                       & 81.05& 26.71& 88.62& 64.44& 85.32& 81.74& 94.97& \underline{68.90}& 88.56& 75.85\\
  \midrule
  TransAttUNet\cite{chen2022transattunet}        & 80.82             & 19.92             & \underline{89.62} & 60.41          & \underline{85.91} & 81.94          & 95.09          & 62.68          & 89.86          & \textbf{81.02} \\
  \rowcolor[RGB]{217,217,217} \textbf{TransAttUNet-EDG (Ours)}  & \underline{82.15$\pm$0.31}$_{\color{red}{+1.33}}$ & 16.98$\pm$1.15$_{\color{red}{-2.94}}$& 89.13$\pm$0.48$_{\color{red}{-0.49}}$ & \underline{67.12$\pm$1.68}$_{\color{red}{+6.71}}$& 85.89$\pm$0.42$_{\color{red}{-0.02}}$ & 82.16$\pm$0.83$_{\color{red}{+0.22}}$& \underline{95.26$\pm$0.19}$_{\color{red}{+0.17}}$ & 67.02$\pm$1.35$_{\color{red}{+4.34}}$& 90.93$\pm$0.37$_{\color{red}{+1.07}}$ & 79.95$\pm$1.04$_{\color{red}{-1.07}}$ \\
  \midrule
  U-Net \cite{ronneberger2015u}                   & 79.98             & 25.91             & 89.09          & 62.47          & 84.93          & 80.07          & 93.97          & 60.09          & 91.08& 78.12 \\
  \rowcolor[RGB]{217,217,217} \textbf{U-Net-EDG (Ours)}  & \textbf{82.96$\pm$0.27}$_{\color{red}{+2.98}}$ & \textbf{14.52$\pm$0.96}$_{\color{red}{-11.39}}$ & \textbf{89.74$\pm$0.38}$_{\color{red}{+0.65}}$  & 64.69$\pm$1.56$_{\color{red}{+2.22}}$& \textbf{86.97$\pm$0.34}$_{\color{red}{+2.04}}$ & \textbf{84.93$\pm$0.67}$_{\color{red}{+4.86}}$ & \textbf{95.51$\pm$0.18}$_{\color{red}{+1.54}}$  & \textbf{69.73$\pm$1.21}$_{\color{red}{+9.64}}$ & \underline{91.47$\pm$0.32}$_{\color{red}{+0.39}}$& \underline{80.82$\pm$0.94}$_{\color{red}{+2.70}}$ \\
  \bottomrule
  \end{tabular}
  }
\end{table*}

\begin{table*}[t]
  \footnotesize
  \centering
  \caption{Performance comparison on the ACDC dataset. Metrics include average DSC (\%) and HD95 (mm) across three cardiac structures, along with per-class DSC. All models were re-implemented and trained under identical experimental settings. \textbf{Bold} indicates the best result; \underline{underlined} denotes the second best. The numbers in red indicate absolute differences (in percentage points for DSC and millimeters for HD95) relative to the original TransAttUNet and U-Net baselines.}
  \label{tab:med_acdc}
  \begin{tabular}{l|cc|ccc}
      \toprule
      \textbf{Model} & \textbf{Avg. DSC $\uparrow$} & \textbf{HD95 $\downarrow$} & \textbf{RV $\uparrow$} & \textbf{Myo $\uparrow$} & \textbf{LV $\uparrow$} \\
      \midrule
      UNext \cite{valanarasu2022unext}               & 86.95             & 2.08             & 85.05             & 84.41             & 91.38             \\
 UNet++\cite{zhou2018unetplusplus}& 91.42& 1.12& 90.68& 89.26&94.33\\
      AttnUNet \cite{oktay2018attnunet}              & 91.45             & 1.11& \underline{90.94}& 89.23& 94.21\\
      MANet \cite{fan2020manet}                      & 91.35& 1.14& 90.67& 89.28& 94.10\\
      \midrule
       TransAttUNet \cite{chen2022transattunet}       & 91.12             & 1.28             & 90.11             & 89.11             & 94.14 \\
        \rowcolor[RGB]{217,217,217} \textbf{TransAttUNet-EDG (Ours)}        & \textbf{91.64$\pm$0.28}$_{\textcolor{red}{+0.52}}$ & \underline{1.09$\pm$0.08}$_{\textcolor{red}{-0.19}}$& \textbf{91.09$\pm$0.41}$_{\textcolor{red}{+0.98}}$& \textbf{89.49$\pm$0.35}$_{\textcolor{red}{+0.38}}$& \textbf{94.35$\pm$0.19}$_{\textcolor{red}{+0.21}}$\\
      \midrule
      U-Net \cite{ronneberger2015u}                   & 90.91             & 1.21             & 89.88             & 89.05             & 93.81             \\
      \rowcolor[RGB]{217,217,217} \textbf{U-Net-EDG (Ours)}                       & \underline{91.49$\pm$0.24}$_{\textcolor{red}{+0.58}}$& \textbf{1.07$\pm$0.07}$_{\textcolor{red}{-0.14}}$ & 90.89$\pm$0.37$_{\textcolor{red}{+1.01}}$& \underline{89.29$\pm$0.31}$_{\textcolor{red}{+0.24}}$& \underline{94.29$\pm$0.16}$_{\textcolor{red}{+0.48}}$\\
      \bottomrule
  \end{tabular}
\end{table*}

\begin{figure*}[t]
    \centering
    \includegraphics[width=1.0\linewidth]{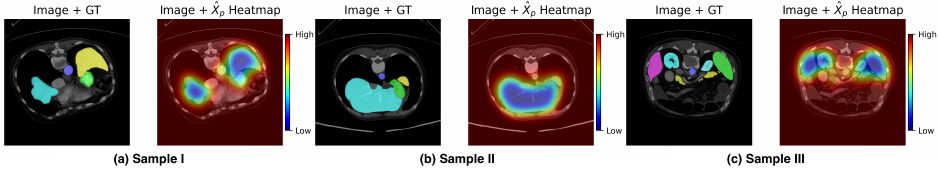}
    \caption{Visual analysis of the per-position entropy ($\hat{\mathbf{X}}_{\text{p}}$, Eq.~\eqref{eq:ep}) on the Synapse dataset. The entropy maps are extracted from the highest-resolution decoder layer and overlaid as heatmaps on the input images. Red indicates high entropy (low certainty); blue indicates low entropy (high certainty).}
    \label{fig:entropy_vis_med}
\end{figure*}

\begin{table}[t]
\centering
\caption{Comparison of computational efficiency among different medical image segmentation methods.}
\label{tab:med_efficiency}
\resizebox{0.5\textwidth}{!}{
\begin{tabular}{lcccc}
\toprule
Methods & Params (M) & MACs (G) & CPU time (ms) & GPU time (ms) \\
\midrule
TransAttUNet & \underline{25.97} & 68.01 & 67.97 & 4.32 \\
\rowcolor[RGB]{217,217,217} \textbf{TransAttUNet-EDG (Ours)} & \textbf{14.39} & \textbf{14.14} & \textbf{46.59} & 4.01 \\
\midrule
U-Net & 32.10 & 44.59 & \underline{54.63} & \textbf{3.22} \\
\rowcolor[RGB]{217,217,217} \textbf{U-Net-EDG (Ours)} & 34.56 & \underline{43.09} & 55.47 & \underline{3.59} \\
\bottomrule
\end{tabular}
}
\end{table}

\begin{figure*}[t]
  \centering
  \includegraphics[width=1.0\textwidth]{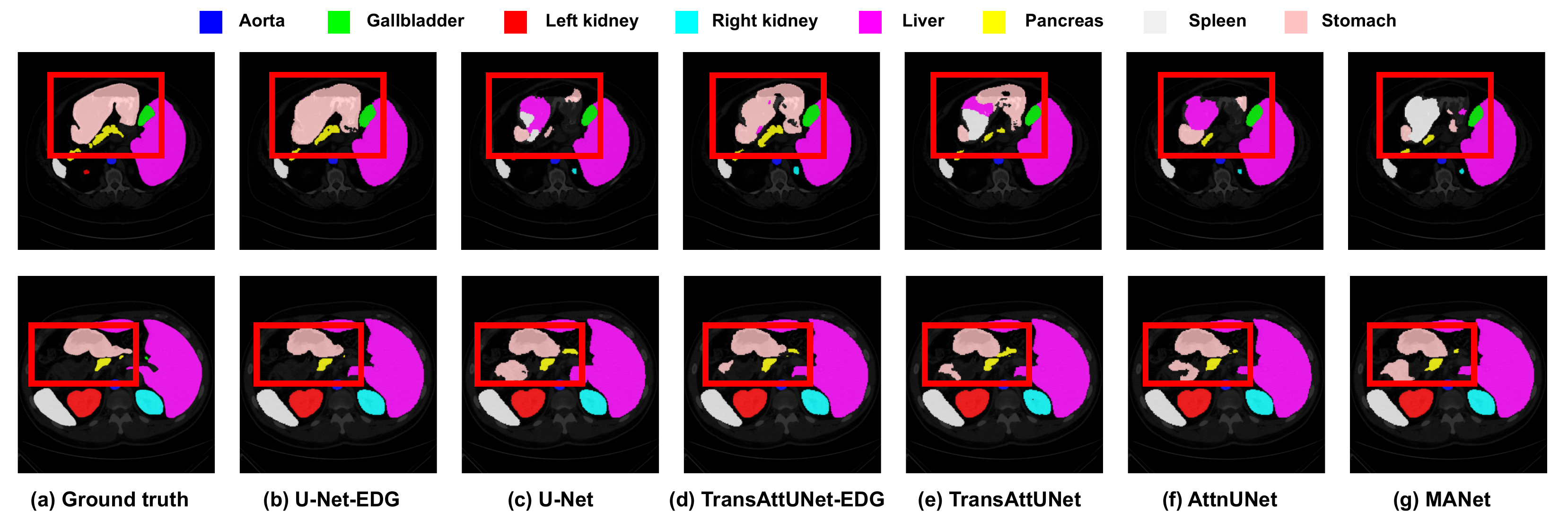}
  \caption{Qualitative comparison of different methods on the Synapse dataset. From left to right: (a) ground truth, (b) U-Net-EDG (ours), (c) U-Net~\cite{ronneberger2015u}, (d) TransAttUNet-EDG (ours), (e) TransAttUNet~\cite{chen2022transattunet}, (f) AttnUNet~\cite{oktay2018attnunet}, and (g) MANet~\cite{fan2020manet}. The red rectangles highlight specific regions of interest to facilitate the comparison of segmentation details. All samples were generated using the checkpoints with the best validation performance.}
  \label{fig:med_samples}
\end{figure*}

\begin{table*}[t]
    \centering
    \setlength{\tabcolsep}{5pt}
    \caption{Quantitative comparison of cloud removal performance between our proposed PMAA-EDG and existing models on two cloud removal benchmarks: Sen2\_MTC\_Old and Sen2\_MTC\_New. The best results are \textbf{bolded}, and the second-best results are \underline{underlined}. The numbers in red denote absolute differences relative to the PMAA \cite{pmaa} baseline (dB for PSNR; unitless for SSIM, FID, LPIPS).}
    \label{tab:main_cloud_removal}
    \resizebox{1.0\textwidth}{!}{
    \begin{tabular}{l|cccc|cccc}
    \toprule
    \multirow{2}{*}{\textbf{Method}} & \multicolumn{4}{c|}{\textbf{Sen2\_MTC\_Old}} & \multicolumn{4}{c}{\textbf{Sen2\_MTC\_New}} \\
    \cmidrule(r){2-5} \cmidrule(r){6-9}
     & \textbf{PSNR}~$\uparrow$ & \textbf{SSIM}~$\uparrow$ & \textbf{FID}~$\downarrow$ & \textbf{LPIPS}~$\downarrow$ & \textbf{PSNR}~$\uparrow$ & \textbf{SSIM}~$\uparrow$ & \textbf{FID}~$\downarrow$ & \textbf{LPIPS}~$\downarrow$ \\
    \midrule
    MCGAN~\cite{mcgan}              & 21.146 & 0.481 & 166.804 & 0.477 & 17.448 & 0.513 & 147.057 & 0.447 \\
    Pix2Pix~\cite{pix2pix}          & 22.894 & 0.437 & 223.446 & 0.557 & 16.985 & 0.455 & 164.524 & 0.535 \\
    AE~\cite{ae}                    & 23.957 & 0.800 & 169.347 & 0.439 & 15.100 & 0.441 & 206.134 & 0.602 \\
    STNet~\cite{stnet}              & 26.321 & 0.834 & 146.057 & 0.438 & 16.206 & 0.427 & 161.683 & 0.503 \\
    DSen2-CR~\cite{dsen2-cr}        & 26.967 & 0.855 & 123.382 & 0.330 & 16.827 & 0.534 & 140.208 & 0.446 \\
    STGAN~\cite{stgan}              & 26.186 & 0.734 & 150.562 & 0.388 & 18.152 & 0.587 & 182.150 & 0.513 \\
    CTGAN~\cite{ctgan}              & 26.264 & 0.808 & 192.270 & 0.472 & 18.308 & 0.609 & 128.704 & 0.384 \\
    CR-TS-Net~\cite{cr-ts-net}      & 26.900 & 0.857 & 121.447 & 0.325 & 18.585 & 0.615 &  96.364 & 0.342 \\
    UnCRtainTS~\cite{uncrtaints}    & 26.417 & 0.837 & 130.875 & 0.400 & \underline{18.770} & \underline{0.631} &  93.509 & 0.333 \\
    DDPM-CR~\cite{ddpm-cr}          & 27.060 & 0.854 & \textbf{110.919} & \underline{0.320} & 18.742 & 0.614 & \underline{94.401} & \textbf{0.329} \\
    \midrule
    PMAA~\cite{pmaa}                & \underline{27.377} & \underline{0.861} & 120.393 & 0.367 & 18.369 & 0.614 & 118.214 & 0.392 \\
    \rowcolor[RGB]{217,217,217}
    \textbf{PMAA-EDG (Ours)}        & \textbf{28.095$\pm$0.182}$_{\color{red}{+0.718}}$ & \textbf{0.877$\pm$0.005}$_{\color{red}{+0.016}}$ & \underline{111.843$\pm$2.14}$_{\color{red}{-8.550}}$ & \textbf{0.291$\pm$0.008}$_{\color{red}{-0.076}}$ & \textbf{19.157$\pm$0.213}$_{\color{red}{+0.788}}$ & \textbf{0.702$\pm$0.006}$_{\color{red}{+0.088}}$ & \textbf{92.315$\pm$2.584}$_{\color{red}{-25.899}}$ & \underline{0.332$\pm$0.009}$_{\color{red}{-0.060}}$ \\
    \bottomrule
    \end{tabular}
    }
\end{table*}

\begin{figure*}[t]
  \centering
  \includegraphics[width=1.0\textwidth]{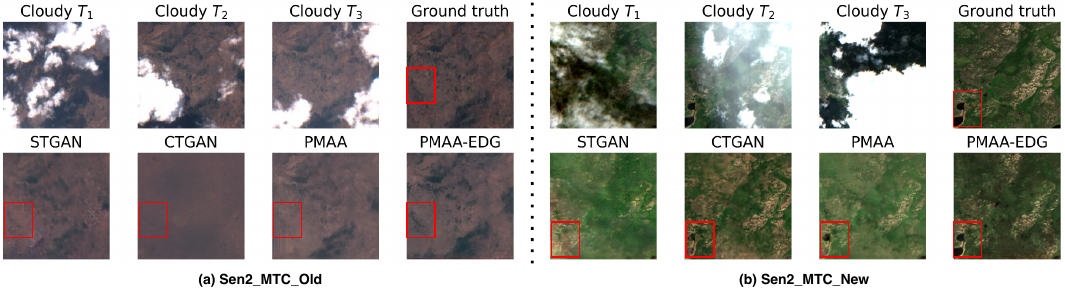}
  \caption{
       Visual comparison of cloud-removal results produced by different methods on two benchmarks: (a) Sen2\_MTC\_Old and (b) Sen2\_MTC\_New. The compared methods include STGAN~\cite{stgan}, CTGAN~\cite{ctgan}, PMAA~\cite{pmaa}, and PMAA-EDG (ours), alongside the cloudy input images ($T_1$, $T_2$, $T_3$) and the corresponding cloud-free ground-truth images. The red rectangles are used to highlight specific regions of interest, allowing for a clearer visual comparison of different methods.}
  \label{fig:main_cloud_removal}
\end{figure*}

\begin{table}[t]
\centering
\caption{Comparison of computational efficiency between PMAA and PMAA-EDG.}
\label{tab:pmaa_efficiency}
\resizebox{0.5\textwidth}{!}{
\begin{tabular}{lcccc}
\toprule
Methods & Params (M) & MACs (G) & CPU time (ms) & GPU time (ms) \\
\midrule
PMAA & \textbf{4.47} & \textbf{112.73} & \textbf{234.54} & \textbf{16.64} \\
\rowcolor[RGB]{217,217,217} \textbf{PMAA-EDG (Ours)} & 6.05 & 122.36 & 297.17 & 28.29 \\
\bottomrule
\end{tabular}
}
\end{table}

\subsubsection{Experimental Settings}
\textbf{Datasets.} We evaluated the EDG module on two publicly available benchmark datasets: the Synapse multi-organ segmentation dataset\footnote{\href{https://www.synapse.org/\#!Synapse:syn3193805/wiki/89480}{https://www.synapse.org/\#!Synapse:syn3193805/wiki/89480}} and the Automated Cardiac Diagnosis Challenge (ACDC) dataset~\cite{bernard2018acdc}. For Synapse dataset, we followed the experimental setup in~\cite{chen2021transunet,fan2020manet}. The 50 contrast-enhanced abdominal CT scans in the Synapse dataset were split into 20 for testing and 30 for training, with the latter split into 18 for training and 12 for internal validation. This task targets eight major abdominal organs: aorta, gallbladder, left kidney, right kidney, liver, pancreas, spleen, and stomach. The ACDC dataset contains 150 short-axis cardiac MRI scans, each annotated with three anatomical structures: right ventricle (RV), myocardium (Myo), and left ventricle (LV). The in-plane spatial resolution ranges from 1.37 to 1.68~mm/pixel, and each sequence contains 28--40 slices covering the full or partial cardiac cycle. The dataset is officially divided into 100 training cases and 50 testing cases.

\noindent \textbf{Training and evaluation.} To ensure comparability with prior work, all input slices were resampled to a uniform spatial resolution of $224\times 224$. Nearest-neighbor interpolation was used for all upsampling operations, and max pooling was employed for spatial downsampling. Model training followed the procedure in~\cite{chen2021transunet}. We used the AdamW optimizer with cosine learning rate decay and trained each model for 400 epochs with a batch size of 8. The initial learning rate was set to $2\times 10^{-4}$, and the weight decay was fixed at $1\times 10^{-4}$. The model was trained by minimizing a joint objective $\mathcal{L} = 0.6 \mathcal{L}_{\text{Dice}} + 0.4 \mathcal{L}_{\text{CE}}$, where $\mathcal{L}_{\text{Dice}}$ denotes the Dice loss and $\mathcal{L}_{\text{CE}}$ denotes the cross-entropy loss. All models were trained from scratch.

\subsubsection{Comparison of Fusion Methods}
\label{sec:comp_fusion}

We first conducted a controlled comparison of fusion methods on the Synapse dataset. We replaced the original fusion module in U-Net~\cite{ronneberger2015u} with the proposed FDG and EDG modules as well as several representative fusion baselines, while keeping the rest of the architecture unchanged.

As shown in Table~\ref{tab:fusion_comparison}, both the FDG and EDG modules outperformed all existing fusion methods in terms of both DSC and HD95. Among the baselines, cross-attention yielded the lowest DSC (76.60\%), likely because its heavy parameterization is prone to overfitting on the moderately sized Synapse training set. The FDG module surpassed the best-performing baseline (concatenation, 79.98\% DSC) by 1.47 percentage points, suggesting that the difference-driven paradigm captures more informative fusion cues than correlation-based approaches. The EDG module achieved the highest scores across all metrics (82.96\% vs.\ 81.45\% DSC and 14.52 vs.\ 16.93 HD95), suggesting that quantifying representational certainty via information entropy provides a more semantically meaningful comparison signal than direct feature differencing. As shown in Table~\ref{tab:fusion_comparison_unet_pmaa}, the EDG and FDG modules had comparable computational costs, both substantially lighter than cross-attention, which nearly doubled the MACs.

\noindent\textbf{Remark:} Across not only the medical image segmentation task but also the other tasks considered in this work, the FDG-based models consistently outperformed existing methods, demonstrating the effectiveness of the difference-based attention mechanism. Since the EDG module achieved further improvements, we report only the EDG results in the following sections for clarity.

\subsubsection{Visual Analysis of Entropy-Based Certainty}
\label{sec:vis_entropy_med}

To provide qualitative insight into what the EDG module learns, we visualized the per-position entropy $\hat{\mathbf{X}}_{\text{p}}$ (Eq.~\eqref{eq:ep}) from the highest-resolution decoder layer. Since $\hat{\mathbf{X}}_{\text{p}}\in\mathbb{R}^{1\times P'}$ can be reshaped into a spatial map, we overlaid it as a heatmap on the input image.

As illustrated in Fig.~\ref{fig:entropy_vis_med}, the learned entropy maps exhibit a clear spatial structure: low-entropy regions (blue) correspond to organ interiors where both the encoder and decoder features tend to agree, while high-entropy regions (red) concentrate at organ boundaries and low-contrast soft-tissue interfaces. This pattern is consistent with the intuition that boundary regions present greater ambiguity for feature fusion, as the encoder preserves fine-grained edge details while the decoder carries coarser semantic context. By producing higher entropy at these locations, the model learns to rely more heavily on whichever stream is locally more certain, which may contribute to the improved boundary delineation observed in the quantitative results.

\subsubsection{Quantitative Results}
As reported in Table~\ref{tab:med_synapse}, the EDG module yielded consistent performance gains on both backbone architectures. On the CNN-based U-Net, integrating the EDG module boosted the average DSC from 79.98\% to 82.96\% ($+2.98$ pp) and reduced HD95 by nearly 44\% (from 25.91~mm to 14.52~mm), achieving the best performance among all compared methods. On the Transformer-based TransAttUNet, the EDG-enhanced variant improved DSC by 1.33 pp (80.82\% $\rightarrow$ 82.15\%) and decreased HD95 by 2.94~mm. Notably, the purely convolutional U-Net-EDG surpassed TransAttUNet-EDG despite lacking explicit long-range dependency modeling, suggesting that the entropy-based gating provides complementary benefits to those offered by self-attention. On the ACDC dataset (Table~\ref{tab:med_acdc}), both variants also showed improvements, with TransAttUNet-EDG and U-Net-EDG reaching 91.64\% and 91.49\% mean DSC, respectively. The gains on ACDC ($+0.52$ and $+0.58$ pp) were smaller than on Synapse, which we attribute to the already high baseline performance on this dataset leaving limited room for improvement.

Beyond segmentation accuracy, we evaluated computational efficiency (Table~\ref{tab:med_efficiency}). Adding the EDG module to U-Net introduced negligible overhead in both parameters and inference time. For TransAttUNet, the EDG-enhanced variant was substantially more compact (14.39M vs.\ 25.97M parameters, 14.14G vs.\ 68.01G MACs). This reduction stems from replacing the original heavy fusion layers in TransAttUNet with the lightweight EDG module, which simultaneously removes redundant cross-attention projections while providing more effective gating.

\subsubsection{Qualitative Results}
Fig.~\ref{fig:med_samples} shows representative segmentation results on the Synapse dataset. In the first row, baselines such as U-Net and TransAttUNet failed to separate adjacent organs (e.g., spleen and stomach), producing merged predictions at their shared boundary. Both U-Net-EDG and TransAttUNet-EDG correctly delineated these boundaries, suggesting that the difference-based gating helps resolve ambiguity in spatially adjacent structures. In the second row, most methods preserved the overall organ shapes, but U-Net-EDG produced noticeably fewer false positives for the stomach and pancreas, two classes that are particularly challenging due to their irregular geometry and low contrast with surrounding tissue.

\subsection{Cloud Removal}
\label{sec:cloud_removal}
\subsubsection{Experimental Settings}
\textbf{Datasets.} We evaluated our method on two benchmark datasets derived from Sentinel-2 imagery: Sen2\_MTC\_Old~\cite{pmaa} and Sen2\_MTC\_New~\cite{ctgan}. The Sen2\_MTC\_Old dataset contains 945 tiles with 3{,}130 multi-temporal image groups. Each group consists of three cloudy acquisitions paired with a corresponding cloud-free reference image. The images have a spatial resolution of $256 \times 256$ pixels across four spectral bands (RGB and near-infrared), with pixel intensities quantized to an 8-bit dynamic range of $[0, 255]$. In our experiments, we partitioned this dataset into training, validation, and test sets with an 8:1:1 ratio following~\cite{pmaa}. The Sen2\_MTC\_New dataset is constructed under a similar multi-temporal paradigm, containing approximately 50 non-overlapping tiles with roughly 70 image groups per tile. While maintaining the same data structure and spectral bands as Sen2\_MTC\_Old, Sen2\_MTC\_New exhibits higher radiometric resolution, with pixel values in a 16-bit range of $[0, 10{,}000]$. It also provides cloud-free reference imagery of superior quality with significantly fewer artifacts. For this dataset, we adopted the official 7:1:2 split for training, validation, and testing, respectively.

\noindent \textbf{Training and evaluation.} To ensure a fair comparison, we conducted all experiments following the protocols established in prior works~\cite{pmaa,ctgan}. Input images were normalized to the range $[-1, 1]$. For multi-temporal processing, we concatenated the three cloudy acquisitions along the channel dimension, yielding a 12-channel input tensor. For the baseline PMAA model, we adopted the network hyperparameters specified in~\cite{pmaa}. All models were trained for 100 epochs using the AdamW optimizer~\cite{adamw} with a batch size of 4. The learning rate was initialized at $5 \times 10^{-4}$, with a weight decay of $1 \times 10^{-5}$, and was modulated using a cosine annealing schedule. To prevent overfitting, we selected the checkpoint with the highest structural similarity index (SSIM) on the validation set for final testing. For quantitative evaluation, we report peak signal-to-noise ratio (PSNR) and SSIM to assess pixel-wise reconstruction fidelity. In addition, we use the Learned Perceptual Image Patch Similarity (LPIPS) metric~\cite{lpips} to measure perceptual quality and the Fréchet Inception Distance (FID)~\cite{fid} to evaluate distributional alignment with real cloud-free imagery.

\subsubsection{Quantitative Results}
Table~\ref{tab:main_cloud_removal} presents the quantitative comparison on the Sen2\_MTC\_Old and Sen2\_MTC\_New datasets. On Sen2\_MTC\_Old, PMAA-EDG achieved the best PSNR (28.095~dB, $+0.718$ over PMAA), SSIM (0.877), and LPIPS (0.291) among all compared methods, while its FID (111.843) ranked second to DDPM-CR (110.919). On the more challenging Sen2\_MTC\_New dataset, which features higher radiometric resolution, PMAA-EDG obtained the highest PSNR (19.157~dB, $+0.788$ over PMAA), SSIM (0.702), and FID (92.315), while its LPIPS (0.332) was comparable to the best result of DDPM-CR (0.329). Notably, the PSNR gain over the strongest non-PMAA competitor (UnCRtainTS, 18.770~dB) was 0.387~dB on Sen2\_MTC\_New, and the SSIM improvement was more pronounced (0.702 vs.\ 0.631). These results suggest that the entropy-based gating enables more effective utilization of multi-temporal information than fixed attention mechanisms. As shown in Table~\ref{tab:pmaa_efficiency}, the EDG module introduced a modest increase in parameters (4.47M $\rightarrow$ 6.05M) and inference time, representing a favorable accuracy--efficiency trade-off given the consistent improvements across both benchmarks.

\subsubsection{Qualitative Results}
Fig.~\ref{fig:main_cloud_removal} shows representative reconstruction results on both benchmarks. On Sen2\_MTC\_Old (Fig.~\ref{fig:main_cloud_removal}(a)), the scene contains arid terrain with complex textures under heavy cloud cover. STGAN and CTGAN produced overly smooth outputs, losing most ground details. PMAA recovered more structure but its output remained hazy with visible loss of high-frequency details (red box). PMAA-EDG reconstructed finer terrain textures, consistent with its lower LPIPS score (0.291 vs.\ 0.367 for PMAA). On Sen2\_MTC\_New (Fig.~\ref{fig:main_cloud_removal}(b)), the scene contains a forest and a lake adjacent to a large invalid-data region (black area), posing a challenge for structural preservation. STGAN and CTGAN produced severe distortions, and PMAA failed to maintain the correct shoreline due to interference from invalid pixels. PMAA-EDG accurately restored the lake boundary, aligning with its substantial SSIM improvement on this dataset (0.702 vs.\ 0.614 for PMAA).

% \begin{figure*}[h]
%   \centering
%   \includegraphics[width=1.0\textwidth]{imgs/spectrogram_report.pdf}
%   \caption{
%        Comparison of spectrograms between the ground-truth audio and the separated results produced by different models on the EchoSet dataset. The figure shows the ground truth, the outputs of TDANet~\cite{li2022efficient} and TIGER~\cite{xu2025tigertimefrequencyinterleavedgain}, and the outputs of their EDG-enhanced variants, TDANet-EDG and TIGER-EDG. The red rectangles highlight specific time-frequency regions of interest, illustrating where our EDG-enhanced models achieve clearer signal separation and fewer artifacts compared to the baseline models.
%     }
%   \label{fig:spectrogram_report}
% \end{figure*}

\subsection{Speech Separation}
\label{sec:speech_sep}
\subsubsection{Experimental Settings}
\textbf{Datasets.} We conducted experiments on two widely used speech separation benchmarks: LRS2-2Mix~\cite{li2022efficient} and EchoSet~\cite{xu2025tigertimefrequencyinterleavedgain}. The LRS2-2Mix dataset was constructed from the LRS2 corpus~\cite{afouras2018deep}, which consists of thousands of video clips collected from the BBC. In contrast to traditional corpora such as WSJ0~\cite{garofolo2007csr} and LibriSpeech~\cite{panayotov2015librispeech}, LRS2 contains recordings from real-world videos with non-stationary background noise and variable reverberation, thus providing a more realistic testbed than clean studio recordings. The data generation protocol follows that of WSJ0-2Mix~\cite{hershey2016deep}, resulting in 20{,}000 training, 5{,}000 validation, and 3{,}000 test mixtures, each 6 seconds in duration and sampled at 16~kHz. The EchoSet dataset is used to assess performance in more complex acoustic environments. EchoSet is constructed using the SonicSim simulator~\cite{lisonicsim} and Matterport3D~\cite{chang2017matterport3d}, enabling accurate simulation of real-world indoor acoustic scenes with both reverberation and noise. It consists of 20{,}268 training, 4{,}604 validation, and 2{,}650 test mixtures, each 2 seconds in duration and sampled at 16~kHz.

\noindent \textbf{Training and evaluation.} To ensure fairness, TIGER~\cite{xu2025tigertimefrequencyinterleavedgain}, TDANet~\cite{li2022efficient}, and their EDG-enhanced variants were trained under identical hyperparameter configurations and optimization schemes. All models were trained for 500 epochs with a batch size of 1. We used the Adam optimizer~\cite{adam} with an initial learning rate of $1\times 10^{-3}$, and optimized the models to maximize the scale-invariant signal-to-noise ratio (SI-SNR)~\cite{le2019sdr}. The learning rate was halved if no performance improvement was observed for 15 consecutive epochs, and training was terminated early if there was no improvement for 30 consecutive epochs. To prevent gradient explosion, we applied gradient clipping with a maximum $\ell_2$ norm of 5. In all experiments, we evaluated the separated audio using scale-invariant signal-to-distortion ratio improvement (SI-SDRi)~\cite{le2019sdr} and signal-to-distortion ratio improvement (SDRi)~\cite{vincent2006performance}.

\begin{table}[t]
    \centering
    \caption{Comparison of different methods on the LRS2-2Mix and EchoSet datasets in terms of SDRi and SI-SDRi (in dB). \textbf{Bold} indicates the best result, while \underline{underlined} denotes the second best. The numbers in red denote improvements (in dB) relative to the original models.}
    \label{tab:speech_separation}
    \renewcommand{\arraystretch}{1.2}
    \resizebox{0.5\textwidth}{!}{
    \begin{tabular}{c|cc|cc}
    \toprule
    \multirow{2}{*}{\textbf{Methods}} & \multicolumn{2}{c|}{\textbf{LRS2-2Mix}} & \multicolumn{2}{c}{\textbf{EchoSet}} \\
                                     & \textbf{SDRi} $\uparrow$ & \textbf{SI-SDRi} $\uparrow$ & \textbf{SDRi} $\uparrow$ & \textbf{SI-SDRi} $\uparrow$ \\
    \midrule
    Conv-TasNet \cite{luo2019conv}         & 11.0 & 10.6 & 7.7 & 6.9 \\
    DualPathRNN \cite{luo2020dual}               & 13.0 & 12.7 & 5.9 & 5.1 \\
    SudoRM-RF1.0x \cite{tzinis2020sudo}          & 11.4 & 11.0 & 7.7 & 6.8 \\
    A-FRCNN-16 \cite{hu2021speech}                & 13.3 & 13.0 & 9.6 & 8.8 \\
    BSRNN \cite{luo2023music}                      & 14.4 & 14.1 & 12.8 & 12.2 \\
    TF-GridNet \cite{wang2023tf}               & \underline{15.7} & \underline{15.4} & \underline{13.7} & \underline{12.9} \\ \midrule
    TDANet \cite{li2022efficient}                     & 14.5 & 14.2 & 10.1 & 9.2 \\
    \rowcolor[RGB]{217,217,217}  \textbf{TDANet-EDG (Ours)}                    & 15.3$\pm$0.12$_{\textcolor{red}{+0.8}}$ & 15.0$\pm$0.11$_{\textcolor{red}{+0.8}}$ & 12.7$\pm$0.19$_{\textcolor{red}{+2.6}}$ & 11.5$\pm$0.17$_{\textcolor{red}{+2.3}}$ \\ \midrule
    TIGER (large) \cite{xu2025tigertimefrequencyinterleavedgain}              & 15.3 & 15.1 & 14.2 & 13.7 \\ 
    \rowcolor[RGB]{217,217,217}  \textbf{TIGER-EDG (Ours)}                    & \textbf{16.3$\pm$0.09}$_{\textcolor{red}{+1.0}}$ & \textbf{16.0$\pm$0.08}$_{\textcolor{red}{+0.9}}$ & \textbf{15.6$\pm$0.14}$_{\textcolor{red}{+1.4}}$ & \textbf{15.0$\pm$0.13}$_{\textcolor{red}{+1.3}}$ \\
    \bottomrule
    \end{tabular}
    }
\end{table}

\begin{table}[t]
\centering
\caption{Comparison of computational efficiency for TDANet and TIGER architectures.}
\label{tab:speech_efficiency}
\resizebox{0.5\textwidth}{!}{
\begin{tabular}{lcccc}
\toprule
Methods & Params (M) & MACs (G) & CPU time (ms) & GPU time (ms) \\
\midrule
TDANet & \textbf{2.33} & \textbf{9.28} & \textbf{441.430} & \textbf{88.079} \\
\rowcolor[RGB]{217,217,217} \textbf{TDANet-EDG (Ours)} & 2.39 & 9.83 & 456.477 & 101.439 \\
\midrule
TIGER & \textbf{0.82} & \textbf{15.48} & \textbf{593.97} & \textbf{102.56} \\
\rowcolor[RGB]{217,217,217} \textbf{TIGER-EDG (Ours)} & 0.92 & 17.24 & 629.62 & 133.18 \\
\bottomrule
\end{tabular}
}
\end{table}

\subsubsection{Quantitative Results}
We investigated the effect of integrating the EDG module into two state-of-the-art U-Net-based architectures, TDANet~\cite{li2022efficient} and TIGER~\cite{xu2025tigertimefrequencyinterleavedgain}. By replacing their fusion layers with the EDG module, we obtained TDANet-EDG and TIGER-EDG, and conducted comparative evaluations against various baselines. Table~\ref{tab:speech_separation} summarizes the performance of the proposed methods. Integrating the EDG module yielded consistent performance gains across different backbones. For TDANet, the EDG-enhanced model improved SDRi / SI-SDRi by 0.8 / 0.8~dB on the LRS2-2Mix test set. On the acoustically more complex EchoSet dataset, the performance gain reached 2.6 / 2.3~dB in SDRi / SI-SDRi, suggesting that the entropy-based gating is particularly beneficial when the acoustic conditions introduce greater feature ambiguity. When applied to the TIGER architecture, the EDG module also led to clear improvements. On LRS2-2Mix, TIGER-EDG achieved an SI-SDRi of 16.0~dB, surpassing the baseline TIGER (15.1~dB) by 0.9~dB. On EchoSet, TIGER-EDG attained an SI-SDRi of 15.0~dB ($+1.3$ over TIGER), establishing state-of-the-art performance across all metrics and datasets. These results indicated that the EDG module was an effective component for enhancing feature fusion in speech separation, functioning robustly across different backbone architectures.

Additionally, we evaluated the computational efficiency of these models. As shown in Table~\ref{tab:speech_efficiency}, integrating the EDG module introduced only marginal computational overhead. For instance, modifying the baseline TDANet added 0.06M parameters and 0.55G MACs, while TIGER-EDG required an increase of 0.10M parameters and 1.76G MACs. The corresponding increases in CPU and GPU inference times remained relatively small across both models.

\subsection{Ablation Study}
\label{sec:ablation}

All ablation experiments in this subsection used the EDG module and were conducted on three representative settings: (1) medical image segmentation using U-Net~\cite{ronneberger2015u} on the Synapse dataset; (2) cloud removal using PMAA~\cite{pmaa} on the Sen2\_MTC\_New dataset~\cite{ctgan}; and (3) speech separation using TIGER~\cite{xu2025tigertimefrequencyinterleavedgain} on the EchoSet dataset~\cite{xu2025tigertimefrequencyinterleavedgain}.

\subsubsection{Gating Dimensions}

\begin{table*}[t]
    \centering
    \caption{Comparison of single-stream modulation methods. Selective-attention and cross-attention (Fig.~\ref{fig:fusion_blocks}(c)(d)) modulate only $\mathbf{L}$; EDG-L is our entropy-based single-stream variant. EDG (dual-stream) is shown for reference. Best single-stream results are \underline{underlined}; overall best in bold.}
    \label{tab:dual_modulation}
    \renewcommand{\arraystretch}{1.2}
    \resizebox{0.7\textwidth}{!}{
    \begin{tabular}{l|cc|cc|cc}
        \toprule
        \multirow{2}{*}{\textbf{Variant}} & \multicolumn{2}{c|}{\textbf{Medical Seg. (UNet)}} & \multicolumn{2}{c|}{\textbf{Cloud Removal (PMAA)}} & \multicolumn{2}{c}{\textbf{Speech Sep. (TIGER)}} \\
        & \textbf{Avg. DSC}~$\uparrow$ & \textbf{HD95}~$\downarrow$ & \textbf{PSNR}~$\uparrow$ & \textbf{SSIM}~$\uparrow$ & \textbf{SDRi}~$\uparrow$ & \textbf{SI-SDRi}~$\uparrow$ \\
        \midrule
        Selective-attention (single-stream) & 79.56 & 28.36 & 18.369 & 0.614 & 14.2 & 13.7 \\
        Cross-attention (single-stream) & 76.60 & 32.74 & 17.894 & 0.563 & 12.1 & 11.5 \\
        EDG-L (single-stream) & \underline{80.91} & \underline{18.27} & \underline{18.536} & \underline{0.637} & \underline{14.6} & \underline{14.1} \\
        \midrule
        \rowcolor[RGB]{217,217,217} \textbf{EDG (dual-stream)} & \textbf{82.96} & \textbf{14.52} & \textbf{19.157} & \textbf{0.702} & \textbf{15.6} & \textbf{15.0} \\
        \bottomrule
    \end{tabular}
    }
\end{table*}

\begin{figure}[t]
  \centering
  \includegraphics[width=0.45\textwidth]{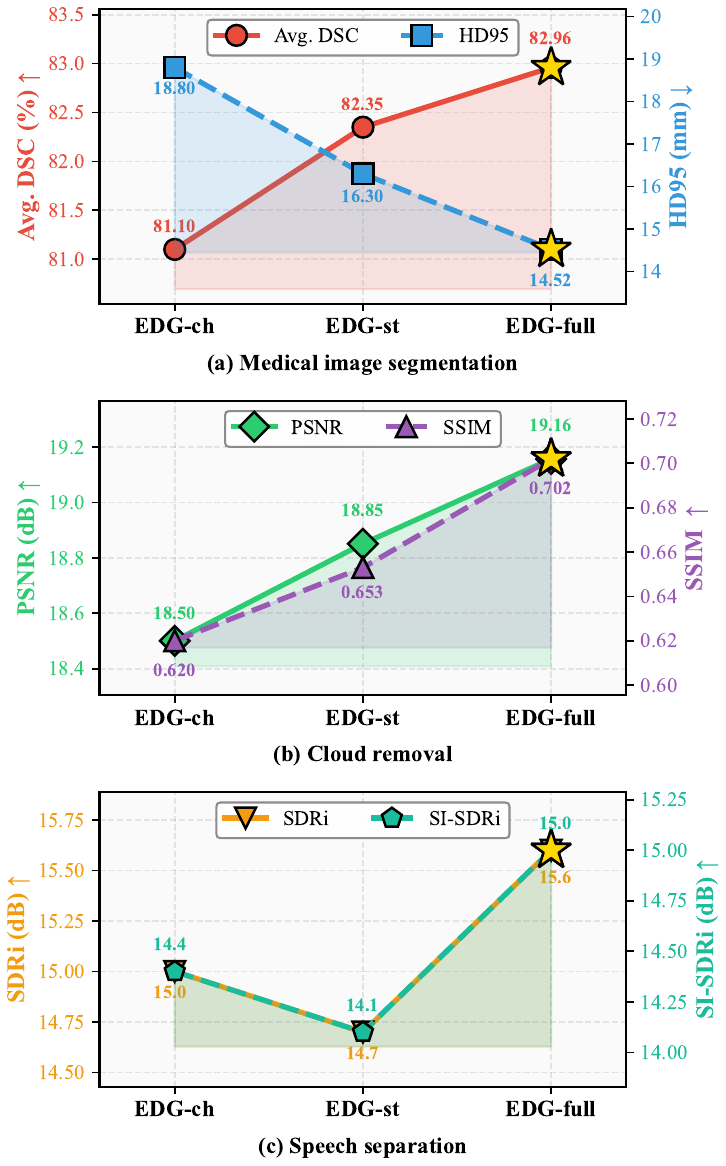}
  \caption{
       Ablation of channel-only (EDG-ch), spatiotemporal-only (EDG-st), and joint channel–spatiotemporal (EDG-full) gating in the EDG module across vision and speech tasks.
    }
  \label{fig:gating_ablation}
\end{figure}

\begin{table*}[t]
    \centering
    \caption{Ablation study on the aggregation function used to derive compressed features in the FE block. All variants share the same GG block (Eqs.~\eqref{eq:kl_general}--\eqref{eq:gate_norm}) and dual-stream modulation; only the aggregation operation that converts the intermediate features $\mathbf{Z}_{\text{n}}^{\mathbf{X}}$, $\mathbf{Z}_{\text{p}}^{\mathbf{X}}$ into the compressed features $\hat{\mathbf{X}}_{\text{n}}$, $\hat{\mathbf{X}}_{\text{p}}$ is varied. \textbf{Bold} indicates the best result.}
    \label{tab:agg_ablation}
    \renewcommand{\arraystretch}{1.2}
    \resizebox{0.8\textwidth}{!}{
    \begin{tabular}{l|cc|cc|cc}
        \toprule
        \multirow{2}{*}{\textbf{Aggregation Function}} & \multicolumn{2}{c|}{\textbf{Medical Seg. (UNet)}} & \multicolumn{2}{c|}{\textbf{Cloud Removal (PMAA)}} & \multicolumn{2}{c}{\textbf{Speech Sep. (TIGER)}} \\
        & \textbf{Avg. DSC}~$\uparrow$ & \textbf{HD95}~$\downarrow$ & \textbf{PSNR}~$\uparrow$ & \textbf{SSIM}~$\uparrow$ & \textbf{SDRi}~$\uparrow$ & \textbf{SI-SDRi}~$\uparrow$ \\
        \midrule
        Mean Pooling (= FDG)         & 81.45$\pm$0.31 & 16.93$\pm$1.08 & 18.723$\pm$0.174 & 0.651$\pm$0.007 & 15.0$\pm$0.13 & 14.4$\pm$0.12 \\
        Variance                     & 81.08$\pm$0.35 & 17.54$\pm$1.22 & 18.691$\pm$0.181 & 0.648$\pm$0.008 & 15.1$\pm$0.15 & 14.5$\pm$0.14 \\
        L2 Norm                      & 81.67$\pm$0.29 & 16.61$\pm$1.03 & 18.794$\pm$0.168 & 0.658$\pm$0.007 & 14.9$\pm$0.14 & 14.3$\pm$0.13 \\
        Learnable FC                 & 82.04$\pm$0.33 & 15.93$\pm$1.11 & 18.881$\pm$0.192 & 0.672$\pm$0.009 & 15.2$\pm$0.16 & 14.6$\pm$0.15 \\
        \rowcolor[RGB]{217,217,217} \textbf{Shannon Entropy (= EDG)} & \textbf{82.96$\pm$0.27} & \textbf{14.52$\pm$0.96} & \textbf{19.157$\pm$0.213} & \textbf{0.702$\pm$0.006} & \textbf{15.6$\pm$0.14} & \textbf{15.0$\pm$0.13} \\

        \bottomrule
    \end{tabular}
    }
\end{table*}

We investigated the importance of modeling certainty along both channel and spatiotemporal dimensions. Concretely, we compared three variants of the GG block: (i) \emph{channel-only gating} (EDG-ch), where we disabled the spatiotemporal gating branch by replacing $\sigma(\Delta_{\text{p}})$ in Eqs.~(\ref{eq:kl_general})–(\ref{eq:kg_general}) with an all-one vector, so that only $\sigma(\Delta_{\text{n}})$ contributed to the gating maps; (ii) \emph{spatiotemporal-only gating} (EDG-st), where we disabled the channel-wise gating branch by replacing $\sigma(\Delta_{\text{n}})$ with an all-one vector, so that only $\sigma(\Delta_{\text{p}})$ was active; and (iii) \emph{full gating} (EDG-full), which corresponded to our complete EDG design and used both $\Delta_{\text{p}}$ and $\Delta_{\text{n}}$ as described in Sec.~\ref{sec:gg_module}.

As illustrated in Fig.~\ref{fig:gating_ablation}, EDG-full consistently outperformed both single-dimensional variants across all benchmarks. Interestingly, the relative importance of the two gating dimensions was task-dependent. On the two vision tasks, EDG-st outperformed EDG-ch (e.g., 82.35\% vs.\ 81.30\% DSC on medical segmentation; 0.653 vs.\ 0.629 SSIM on cloud removal), indicating that spatiotemporal certainty plays a more critical role in image-based fusion where spatial structure is the primary cue. In contrast, on speech separation, EDG-ch outperformed EDG-st (14.7 vs.\ 13.5~dB SI-SDRi), suggesting that channel-wise certainty is more informative for speech signals, where each feature channel encodes distinct spectral characteristics. Despite these task-specific preferences, combining both dimensions in EDG-full consistently yielded the best results, confirming that channel-wise and spatiotemporal certainty provide complementary information.

\subsubsection{Dual-Stream vs.\ Single-Stream Modulation}
\label{sec:dual_modulation}

As discussed in Sec.~\ref{sec:related_works}, existing attention-based fusion methods typically modulate only the local features $\mathbf{L}$ while leaving the global features $\mathbf{G}$ unchanged (cf.\ Fig.~\ref{fig:fusion_blocks}(c) and (d)). To verify whether simultaneously modulating both streams is beneficial, we compared the full EDG module against a single-stream variant (EDG-L), in which only $\mathbf{L}$ is gated while $\mathbf{G}$ passes through without modulation. Concretely, in EDG-L we fix $\tilde{\mathbf{K}}_{G}=\mathbf{1}$ in the final fusion step, so that the upsampled global content feature $\phi(\hat{\mathbf{G}}_{\text{d}})$ is added without attenuation, and only $\hat{\mathbf{L}}_{\text{d}}$ is modulated by $\tilde{\mathbf{K}}_{L}$.

As shown in Table~\ref{tab:dual_modulation}, EDG-L outperformed both selective-attention and cross-attention across all three tasks (e.g., $+1.35$ DSC and $-10.09$ HD95 over selective-attention on medical segmentation; $+0.4$~dB SI-SDRi on speech separation), showing that entropy-based gating is more effective even when applied to only one stream. The full dual-stream EDG module improved over EDG-L by a substantial margin on every benchmark ($+2.05$ DSC, $+0.621$~dB PSNR, $+0.9$~dB SI-SDRi), indicating that simultaneously modulating both $\mathbf{G}$ and $\mathbf{L}$ is highly beneficial for fusion quality.

\subsubsection{Choice of Aggregation Function}
\label{sec:ablation_agg}

A central design of the proposed framework is the aggregation function that converts the intermediate features $\mathbf{Z}_{\text{n}}^{\mathbf{X}}$ and $\mathbf{Z}_{\text{p}}^{\mathbf{X}}$ into the per-channel feature $\hat{\mathbf{X}}_{\text{n}}$ (compressed along the spatiotemporal dimension) and the per-position feature $\hat{\mathbf{X}}_{\text{p}}$ (compressed along the channel dimension). In FDG, this is instantiated as mean pooling (Eqs.~\eqref{eq:fdg_n}--\eqref{eq:fdg_p}); in EDG, it is instantiated as softmax-normalized Shannon entropy (Eqs.~\eqref{eq:en}--\eqref{eq:ep}). To isolate the contribution of the entropy formulation from the broader dual-stream gating framework, we compared it against several alternative aggregation functions while keeping all other components strictly identical. Specifically, we evaluated the following variants:

\begin{itemize}
    \item \textbf{Mean Pooling} (equivalent to FDG): Averages across the complementary dimension, yielding first-order statistics.
    \item \textbf{Variance}: Computes the variance across the complementary dimension, i.e., $\hat{\mathbf{X}}_{\text{n}}(c) = \text{Var}_{q}(\mathbf{Z}_{\text{n}}^{\mathbf{X}}(c,:))$ and $\hat{\mathbf{X}}_{\text{p}}(q) = \text{Var}_{c}(\mathbf{Z}_{\text{p}}^{\mathbf{X}}(:,q))$. This captures second-order spread information but lacks a distributional interpretation.
    \item \textbf{L2 Norm}: Computes the L2 norm across the complementary dimension, i.e., $\hat{\mathbf{X}}_{\text{n}}(c) = \lVert \mathbf{Z}_{\text{n}}^{\mathbf{X}}(c,:) \rVert_2$ and $\hat{\mathbf{X}}_{\text{p}}(q) = \lVert \mathbf{Z}_{\text{p}}^{\mathbf{X}}(:,q) \rVert_2$. This reflects the overall activation magnitude.
    \item \textbf{Learnable FC}: Applies a learnable linear projection to reduce the complementary dimension to a scalar, i.e., $\hat{\mathbf{X}}_{\text{n}} = \mathbf{Z}_{\text{n}}^{\mathbf{X}} \mathbf{w}_{\text{n}}$ and $\hat{\mathbf{X}}_{\text{p}} = \mathbf{w}_{\text{p}}^{\top} \mathbf{Z}_{\text{p}}^{\mathbf{X}}$, where $\mathbf{w}_{\text{n}} \in \mathbb{R}^{Q \times 1}$ and $\mathbf{w}_{\text{p}} \in \mathbb{R}^{C \times 1}$ are learnable weight vectors. This provides maximum flexibility but lacks inductive bias.
    \item \textbf{Shannon Entropy} (equivalent to EDG): Applies softmax normalization followed by Shannon entropy computation, as defined in Eqs.~\eqref{eq:en}--\eqref{eq:ep}.
\end{itemize}

For all non-entropy variants, the difference metrics $\Delta_{\text{n}}$ and $\Delta_{\text{p}}$ were computed as absolute differences following Eqs.~\eqref{eq:deltan_feat}--\eqref{eq:deltap_feat}, consistent with the FDG formulation. For the entropy variant, signed differences were used as in Eqs.~\eqref{eq:dn_ent}--\eqref{eq:dp_ent}, since the sign of the entropy difference carries semantic meaning (indicating which stream is more certain).

As shown in Table~\ref{tab:agg_ablation}, Shannon entropy consistently achieved the best performance across all three tasks, improving over mean pooling (FDG) by 1.51~pp in DSC (14.52 vs.\ 16.93 in HD95), 0.434~dB in PSNR (0.702 vs.\ 0.651 in SSIM), and 0.6~dB in SI-SDRi (15.6 vs.\ 15.0 in SDRi). Among the fixed statistics (Mean, Variance, L2), no single one dominated: L2 Norm performed best on the two vision tasks while Variance was strongest on speech separation. The Learnable FC variant consistently ranked second across all tasks, outperforming all fixed-statistic alternatives, yet still fell short of Shannon entropy on every metric despite having the most degrees of freedom. These results suggest that entropy's advantage stems not from added capacity but from its distributional inductive bias, which provides semantically meaningful signed comparisons that learnable projections do not naturally recover.

\section{Conclusions}
\label{sec:conclusions}

We proposed two lightweight gating modules, FDG and EDG, for encoder--decoder feature fusion in U-Net architectures. Both modules derive adaptive gating maps from the difference between the two feature streams and simultaneously modulate the global and local branches with negligible computational overhead. The key distinction is that the EDG module replaces the absolute feature difference used in the FDG module with a signed entropy difference, enabling the gating signal to reflect the relative representational certainty of each stream. A controlled comparison on medical image segmentation showed that both modules consistently outperformed existing fusion methods. Comprehensive experiments on cloud removal and speech separation confirmed that the EDG module achieved the state-of-the-art results across all three tasks. These results suggest that difference-based gating provides an effective strategy across diverse data modalities.

% use section* for acknowledgment
\ifCLASSOPTIONcompsoc
  % The Computer Society usually uses the plural form
  \section*{Acknowledgments}
\else
  % regular IEEE prefers the singular form
  \section*{Acknowledgment}
\fi

This work was supported by the National Key Research and Development Program of China (No. 2021ZD0200301), the National Natural Science Foundation of China (Nos. U2341228 and 62576187), and the Fundamental and Interdisciplinary Disciplines Breakthrough Plan of the Ministry of Education of China (No. JYB2025XDXM504).

% Can use something like this to put references on a page
% by themselves when using endfloat and the captionsoff option.
\ifCLASSOPTIONcaptionsoff
  \newpage
\fi

% trigger a \newpage just before the given reference
% number - used to balance the columns on the last page
% adjust value as needed - may need to be readjusted if
% the document is modified later
%\IEEEtriggeratref{8}
% The "triggered" command can be changed if desired:
%\IEEEtriggercmd{\enlargethispage{-5in}}

% references section

% can use a bibliography generated by BibTeX as a .bbl file
% BibTeX documentation can be easily obtained at:
% http://mirror.ctan.org/biblio/bibtex/contrib/doc/
% The IEEEtran BibTeX style support page is at:
% http://www.michaelshell.org/tex/ieeetran/bibtex/
\bibliographystyle{IEEEtran}
% argument is your BibTeX string definitions and bibliography database(s)
\bibliography{references}

@inproceedings{li2022efficient,
  title={An efficient encoder-decoder architecture with top-down attention for speech separation},
  author={Li, Kai and Yang, Runxuan and Hu, Xiaolin},
  booktitle={International Conference on Learning Representations},
  year={2022}
}

@inproceedings{xu2025tigertimefrequencyinterleavedgain,
  title={{TIGER}: Time-frequency interleaved gain extraction and reconstruction for efficient speech separation},
  author={Xu, Mohan and Li, Kai and Chen, Guo and Hu, Xiaolin},
  year={2024},
  booktitle={International Conference on Learning Representations}
}

@inproceedings{ronneberger2015u,
  title={{U-Net}: Convolutional networks for biomedical image segmentation},
  author={Ronneberger, Olaf and Fischer, Philipp and Brox, Thomas},
  booktitle={International Conference on Medical Image Computing and Computer-Assisted Intervention},
  pages={234--241},
  year={2015},
  organization={Springer}
}

@inproceedings{zhou2018unet++,
  title={{UNet++}: A nested {U-Net} architecture for medical image segmentation},
  author={Zhou, Zongwei and Rahman Siddiquee, Md Mahfuzur and Tajbakhsh, Nima and Liang, Jianming},
  booktitle={Deep Learning in Medical Image Analysis and Multimodal Learning for Clinical Decision Support},
  pages={3--11},
  year={2018},
  organization={Springer}
}

@inproceedings{oktay2018attnunet,
  author = {Oktay, Ozan and Schlemper, Jo and Le Folgoc, Lo{\"i}c and Lee, Matthew C. H. and Heinrich, Mattias P. and Misawa, Kazunari and Mori, Kensaku and McDonagh, Steven G. and Hammerla, Nils Y. and Kainz, Bernhard and Glocker, Ben and Rueckert, Daniel},
  title = {Attention {U-Net}: learning where to look for the pancreas},
  booktitle = {Medical Imaging with Deep Learning},
  year = {2018}
}

@inproceedings{cai2020maunet,
  title = {{MA-Unet}: an improved version of {Unet} based on multi-scale and attention mechanism for medical image segmentation}, 
  author={Cai, Yutong and Wang, Yong},
  booktitle = {Third International Conference on Electronics and Communication; Network and Computer Technology},
  year = {2020}
}

@article{fan2020manet,
  author = {Fan, Tongle and Wang, Guanglei and Li, Yan and Wang, Hongrui},
  year = {2020},
  pages = {179656--179665},
  title = {{MA-Net}: a multi-scale attention network for liver and tumor segmentation},
  volume = {8},
  journal = {IEEE Access},
  publisher={IEEE}
}

@misc{zhou2018unetplusplus,
  title={{UNet++}: a nested {U-Net} architecture for medical image segmentation}, 
  author={Zhou, Zongwei and Rahman Siddiquee, Md Mahfuzur and Tajbakhsh, Nima and Liang, Jianming},
  year={2018},
  eprint={1807.10165},
  archivePrefix={arXiv}
}

@inproceedings{valanarasu2022unext,
  title={{UNeXt}: {MLP}-based rapid medical image segmentation network}, 
  author={Valanarasu, Jeya Maria Jose and Patel, Vishal M.},
  year={2022},
  booktitle={International Conference on Medical Image Computing and Computer-Assisted Intervention}
}

@article{chen2021transunet,
  author       = {Chen, Jieneng and Lu, Yongyi and Yu, Qihang and Luo, Xiangde and Adeli, Ehsan and Wang, Yan and Lu, Le and Yuille, Alan L. and Zhou, Yuyin},
  title        = {{TransUNet}: {Transformers} make strong encoders for medical image segmentation},
  journal      = {Computing Research Repository},
  volume       = {abs/2102.04306},
  year         = {2021}
}

@inproceedings{cao2021swinunet,
   author = {Cao, Hu and Wang, Yueyue and Chen, Joy and Jiang, Dongsheng and Zhang, Xiaopeng and Tian, Qi and Wang, Manning},
   title = {{Swin-Unet}: {Unet}-like pure {Transformer} for medical image segmentation},
   booktitle = {Proceedings of the European Conference on Computer Vision Workshops},
   year = {2022}
}

@article{bernard2018acdc,
  author={Bernard, Olivier and Lalande, Alain and Zotti, Clement and Cervenansky, Frederick and Yang, Xin and Heng, Pheng-Ann and Cetin, Irem and Lekadir, Karim and Camara, Oscar and Gonzalez Ballester, Miguel Angel and Sanroma, Gerard and Napel, Sandy and Petersen, Steffen and Tziritas, Georgios and Grinias, Elias and Khened, Mahendra and Kollerathu, Varghese Alex and Krishnamurthi, Ganapathy and Roh{\'e}, Marc-Michel and Pennec, Xavier and Sermesant, Maxime and Isensee, Fabian and J{\"a}ger, Paul and Maier-Hein, Klaus H. and Full, Peter M. and Wolf, Ivo and Engelhardt, Sandy and Baumgartner, Christian F. and Koch, Lisa M. and Wolterink, Jelmer M. and I{\v{s}}gum, Ivana and Jang, Yeonggul and Hong, Yoonmi and Patravali, Jay and Jain, Shubham and Humbert, Olivier and Jodoin, Pierre-Marc},
  journal={IEEE Transactions on Medical Imaging}, 
  title={Deep learning techniques for automatic {MRI} cardiac multi-structures segmentation and diagnosis: is the problem solved?}, 
  year={2018},
  volume={37},
  number={11},
  pages={2514--2525},
  publisher={IEEE}
}

@article{chen2022transattunet,
  author={Chen, Bingzhi and Liu, Yishu and Zhang, Zheng and Lu, Guangming and Kong, Adams Wai Kin},
  journal={IEEE Transactions on Emerging Topics in Computational Intelligence}, 
  title={{TransAttUnet}: multi-level attention-guided {U-Net} with {Transformer} for medical image segmentation}, 
  year={2024},
  volume={8},
  number={1},
  pages={55--68},
  publisher={IEEE}
}

@inproceedings{hanDeepPredictiveCoding2018,
  title = {Deep predictive coding network with local recurrent processing for object recognition},
  booktitle = {Advances in Neural Information Processing Systems},
  author = {Han, Kuan and Wen, Haiguang and Zhang, Yizhen and Fu, Di and Culurciello, E. and Liu, Zhongming},
  year = 2018,
  month = may,
}

@inproceedings{adam,
  title={{Adam}: a method for stochastic optimization},
  author={Kingma, Diederik P. and Ba, Jimmy},
  booktitle={International Conference on Learning Representations},
  year={2015}
}

@inproceedings{mcgan,
  title={Filmy cloud removal on satellite imagery with multispectral conditional generative adversarial nets},
  author={Enomoto, Kenji and Sakurada, Ken and Wang, Weimin and Fukui, Hiroshi and Matsuoka, Masashi and Nakamura, Ryosuke and Kawaguchi, Nobuo},
  booktitle={Proceedings of the {IEEE/CVF} Conference on Computer Vision and Pattern Recognition Workshops},
  pages={48--56},
  year={2017}
}

@inproceedings{pix2pix,
  title={Image-to-image translation with conditional adversarial networks},
  author={Isola, Phillip and Zhu, Jun-Yan and Zhou, Tinghui and Efros, Alexei A},
  booktitle={Proceedings of the {IEEE/CVF} Conference on Computer Vision and Pattern Recognition},
  pages={1125--1134},
  year={2017}
}

@inproceedings{ae,
  title={A multi-temporal convolutional autoencoder neural network for cloud removal in remote sensing images},
  author={Sintarasirikulchai, Wassana and Kasetkasem, Teerasit and Isshiki, Tsuyoshi and Chanwimaluang, Thitiporn and Rakwatin, Preesan},
  booktitle={International Conference on Electrical Engineering/Electronics, Computer, Telecommunications and Information Technology},
  pages={360--363},
  year={2018}
}

@article{stnet,
  title={Thick clouds removing from multitemporal {Landsat} images using spatiotemporal neural networks},
  author={Chen, Yang and Weng, Qihao and Tang, Luliang and Zhang, Xia and Bilal, Muhammad and Li, Qingquan},
  journal={IEEE Transactions on Geoscience and Remote Sensing},
  volume={60},
  pages={1--14},
  year={2020},
  publisher={IEEE}
}

@article{dsen2-cr,
  title = {Cloud removal in {Sentinel-2} imagery using a deep residual neural network and {SAR}-optical data fusion},
  journal = {ISPRS Journal of Photogrammetry and Remote Sensing},
  volume = {166},
  pages = {333--346},
  year = {2020},
  author = {Meraner, Andrea and Ebel, Patrick and Zhu, Xiao Xiang and Schmitt, Michael},
  publisher={Elsevier}
}

@inproceedings{stgan,
  title={Cloud removal in satellite images using spatiotemporal generative networks},
  author={Sarukkai, Vishnu and Jain, Anirudh and Uzkent, Burak and Ermon, Stefano},
  booktitle={Winter Conference on Applications of Computer Vision},
  pages={1796--1805},
  year={2020}
}

@inproceedings{ctgan,
  title={{CTGAN}: cloud {Transformer} generative adversarial network},
  author={Huang, Gi-Luen and Wu, Pei-Yuan},
  booktitle={International Conference on Image Processing},
  pages={511--515},
  year={2022},
  organization={IEEE}
}

@article{cr-ts-net,
  title={{SEN12MS-CR-TS}: a remote-sensing data set for multimodal multitemporal cloud removal},
  author={Ebel, Patrick and Xu, Yajin and Schmitt, Michael and Zhu, Xiao Xiang},
  journal={IEEE Transactions on Geoscience and Remote Sensing},
  volume={60},
  pages={1--14},
  year={2022},
  publisher={IEEE}
}

@inproceedings{pmaa,
  title = {{PMAA}: a progressive multi-scale attention autoencoder model for high-performance cloud removal from multi-temporal satellite imagery},
  author = {Zou, Xuechao and Li, Kai and Xing, Junliang and Tao, Pin and Cui, Yachao},
  booktitle = {European Conference on Artificial Intelligence},
  year = {2023},
  pages={3165--3172}
}

@inproceedings{uncrtaints,
  title = {{UnCRtainTS}: uncertainty quantification for cloud removal in optical satellite time series},
  author = {Ebel, Patrick and Garnot, Vivien Sainte Fare and Schmitt, Michael and Wegner, Jan and Zhu, Xiao Xiang},
  booktitle = {Proceedings of the {IEEE/CVF} Conference on Computer Vision and Pattern Recognition Workshops},
  year = {2023},
  pages={2086--2096}
}

@article{ddpm-cr,
  author = {Jing, Ran and Duan, Fuzhou and Lu, Fengxian and Zhang, Miao and Zhao, Wenji},
  title = {Denoising diffusion probabilistic feature-based network for cloud removal in {Sentinel-2} imagery},
  journal = {Remote Sensing},
  volume = {15},
  year = {2023},
  number = {9},
  publisher={MDPI}
}

@article{fid,
  title={The {Fr{\'e}chet} distance between multivariate normal distributions},
  author={Dowson, D. C. and Landau, B. V.},
  journal={Journal of Multivariate Analysis},
  volume={12},
  number={3},
  pages={450--455},
  year={1982},
  publisher={Elsevier}
}

@inproceedings{lpips,
  title={The unreasonable effectiveness of deep features as a perceptual metric},
  author={Zhang, Richard and Isola, Phillip and Efros, Alexei A and Shechtman, Eli and Wang, Oliver},
  booktitle={Proceedings of the {IEEE/CVF} Conference on Computer Vision and Pattern Recognition},
  pages={586--595},
  year={2018}
}

@inproceedings{adamw,
  title={Decoupled weight decay regularization},
  author={Loshchilov, Ilya and Hutter, Frank},
  booktitle={International Conference on Learning Representations},
  pages={1--18},
  year={2018}
}

@article{friston2008hierarchical,
  title={Hierarchical models in the brain},
  author={Friston, Karl},
  journal={PLoS Computational Biology},
  volume={4},
  number={11},
  pages={e1000211},
  year={2008},
  publisher={Public Library of Science San Francisco, USA}
}

@book{pribram2013brain,
  title={Brain and perception: holonomy and structure in figural processing},
  author={Pribram, Karl H},
  year={2013},
  publisher={Psychology Press}
}

@article{groen2017contributions,
  title={Contributions of low- and high-level properties to neural processing of visual scenes in the human brain},
  author={Groen, Iris IA and Silson, Edward H and Baker, Chris I},
  journal={Philosophical Transactions of the Royal Society B: Biological Sciences},
  volume={372},
  number={1714},
  pages={20160102},
  year={2017},
  publisher={The Royal Society}
}

@article{skoe2010auditory,
  title={Auditory brain stem response to complex sounds: a tutorial},
  author={Skoe, Erika and Kraus, Nina},
  journal={Ear and Hearing},
  volume={31},
  number={3},
  pages={302--324},
  year={2010},
  publisher={LWW}
}

@article{azad2024medical,
  title={Medical image segmentation review: the success of {U-Net}},
  author={Azad, Reza and Aghdam, Ehsan Khodapanah and Rauland, Amelie and Jia, Yiwei and Avval, Atlas Haddadi and Bozorgpour, Afshin and Karimijafarbigloo, Sanaz and Cohen, Joseph Paul and Adeli, Ehsan and Merhof, Dorit},
  journal={IEEE Transactions on Pattern Analysis and Machine Intelligence},
  year={2024},
  publisher={IEEE}
}

@article{shen2014effective,
  title={An effective thin cloud removal procedure for visible remote sensing images},
  author={Shen, Huanfeng and Li, Huifang and Qian, Yan and Zhang, Liangpei and Yuan, Qiangqiang},
  journal={ISPRS Journal of Photogrammetry and Remote Sensing},
  volume={96},
  pages={224--235},
  year={2014},
  publisher={Elsevier}
}

@article{chen2023auto,
  title={Auto-encoders in deep learning---a review with new perspectives},
  author={Chen, Shuangshuang and Guo, Wei},
  journal={Mathematics},
  volume={11},
  number={8},
  pages={1777},
  year={2023},
  publisher={MDPI}
}

@article{luo2019conv,
  author={Luo, Yi and Mesgarani, Nima},
  journal={IEEE/ACM Transactions on Audio, Speech, and Language Processing}, 
  title={{Conv-TasNet}: surpassing ideal time-frequency magnitude masking for speech separation}, 
  year={2019},
  volume={27},
  number={8},
  pages={1256--1266},
  publisher={IEEE}
}

@inproceedings{luo2020dual,
  author={Luo, Yi and Chen, Zhuo and Yoshioka, Takuya},
  booktitle={IEEE International Conference on Acoustics, Speech and Signal Processing}, 
  title={Dual-path {RNN}: efficient long sequence modeling for time-domain single-channel speech separation}, 
  year={2020},
  pages={46--50},
  organization={IEEE}
}

@inproceedings{wang2023tf,
  author={Wang, Zhong-Qiu and Cornell, Samuele and Choi, Shukjae and Lee, Younglo and Kim, Byeong-Yeol and Watanabe, Shinji},
  booktitle={IEEE International Conference on Acoustics, Speech and Signal Processing}, 
  title={{TF-GRIDNET}: making time-frequency domain models great again for monaural speaker separation}, 
  year={2023},
  pages={1--5},
  organization={IEEE}
}

@inproceedings{tzinis2020sudo,
  title={{Sudo RM -RF}: efficient networks for universal audio source separation},
  author={Tzinis, Efthymios and Wang, Zhepei and Smaragdis, Paris},
  booktitle={2020 IEEE 30th International Workshop on Machine Learning for Signal Processing},
  pages={1--6},
  year={2020},
  organization={IEEE}
}

@article{hu2021speech,
  title={Speech separation using an asynchronous fully recurrent convolutional neural network},
  author={Hu, Xiaolin and Li, Kai and Zhang, Weiyi and Luo, Yi and Lemercier, Jean-Marie and Gerkmann, Timo},
  journal={Advances in Neural Information Processing Systems},
  volume={34},
  pages={22509--22522},
  year={2021},
  publisher={Curran Associates, Inc.}
}

@misc{hendrycks2016gaussian,
  title={Gaussian error linear units ({GELUs})},
  author={Hendrycks, Dan and Gimpel, Kevin},
  year={2016},
  eprint={1606.08415},
  archivePrefix={arXiv}
}

@inproceedings{ioffe2015batch,
  title={Batch normalization: accelerating deep network training by reducing internal covariate shift},
  author={Ioffe, Sergey and Szegedy, Christian},
  booktitle={Proceedings of the 32nd International Conference on Machine Learning},
  pages={448--456},
  year={2015},
  organization={PMLR}
}

@article{afouras2018deep,
  title={Deep audio-visual speech recognition},
  volume={44},
  number={12},
  journal={IEEE Transactions on Pattern Analysis and Machine Intelligence},
  publisher={Institute of Electrical and Electronics Engineers},
  author={Afouras, Triantafyllos and Chung, Joon Son and Senior, Andrew and Vinyals, Oriol and Zisserman, Andrew},
  year={2022},
  pages={8717--8727}
}

@misc{garofolo2007csr,
  author = {Garofolo, John S. and Graff, David and Baker, Janet M. and Paul, Doug and Pallett, David},
  title = {{CSR-I} ({WSJ0}) complete},
  year = {2007},
  publisher = {Linguistic Data Consortium}
}

@inproceedings{panayotov2015librispeech,
  author={Panayotov, Vassil and Chen, Guoguo and Povey, Daniel and Khudanpur, Sanjeev},
  booktitle={IEEE International Conference on Acoustics, Speech and Signal Processing}, 
  title={{Librispeech}: an {ASR} corpus based on public domain audio books}, 
  year={2015},
  pages={5206--5210},
  organization={IEEE}
}

@inproceedings{hershey2016deep,
  author={Hershey, John R. and Chen, Zhuo and Le Roux, Jonathan and Watanabe, Shinji},
  booktitle={IEEE International Conference on Acoustics, Speech and Signal Processing}, 
  title={Deep clustering: discriminative embeddings for segmentation and separation}, 
  year={2016},
  pages={31--35},
  organization={IEEE}
}

@inproceedings{chang2017matterport3d,
  title={{Matterport3D}: learning from {RGB-D} data in indoor environments},
  author={Chang, Angel and Dai, Angela and Funkhouser, Thomas and Halber, Maciej and Niebner, Matthias and Savva, Manolis and Song, Shuran and Zeng, Andy and Zhang, Yinda},
  booktitle={International Conference on 3D Vision},
  year={2017}
}

@inproceedings{le2019sdr,
  author={Le Roux, Jonathan and Wisdom, Scott and Erdogan, Hakan and Hershey, John R.},
  booktitle={IEEE International Conference on Acoustics, Speech and Signal Processing}, 
  title={{SDR} --- half-baked or well done?}, 
  year={2019},
  pages={626--630},
  organization={IEEE}
}

@article{vincent2006performance,
  author={Vincent, E. and Gribonval, R. and F{\'e}votte, C.},
  journal={IEEE Transactions on Audio, Speech, and Language Processing}, 
  title={Performance measurement in blind audio source separation}, 
  year={2006},
  volume={14},
  number={4},
  pages={1462--1469},
  publisher={IEEE}
}

@article{luo2023music,
  author={Luo, Yi and Yu, Jianwei},
  journal={IEEE/ACM Transactions on Audio, Speech, and Language Processing}, 
  title={Music source separation with band-split {RNN}}, 
  year={2023},
  volume={31},
  pages={1893--1901},
  publisher={IEEE}
}

@article{shannon1948mathematical,
  title={A mathematical theory of communication},
  author={Shannon, Claude E},
  journal={The Bell System Technical Journal},
  volume={27},
  number={3},
  pages={379--423},
  year={1948},
  publisher={Nokia Bell Labs}
}

@inproceedings{zhang2018exfuse,
  title={{Exfuse}: enhancing feature fusion for semantic segmentation},
  author={Zhang, Zhenli and Zhang, Xiangyu and Peng, Chao and Xue, Xiangyang and Sun, Jian},
  booktitle={Proceedings of the European Conference on Computer Vision ({ECCV})},
  pages={269--284},
  year={2018}
}

@inproceedings{lisonicsim,
  title={{SonicSim}: a customizable simulation platform for speech processing in moving sound source scenarios},
  author={Li, Kai and Sang, Wendi and Zeng, Chang and Yang, Runxuan and Chen, Guo and Hu, Xiaolin},
  booktitle={The Thirteenth International Conference on Learning Representations},
  year={2025}
}

@article{diakogiannis2020resunet,
  title={{ResUNet-a}: a deep learning framework for semantic segmentation of remotely sensed data},
  author={Diakogiannis, Foivos I and Waldner, Fran{\c{c}}ois and Caccetta, Peter and Wu, Chen},
  journal={ISPRS Journal of Photogrammetry and Remote Sensing},
  volume={162},
  pages={94--114},
  year={2020},
  publisher={Elsevier}
}

@article{hickok2007corticalspeech,
  title = {The cortical organization of speech processing},
  author = {Hickok, Gregory and Poeppel, David},
  date = {2007-05},
  journaltitle = {Nature Reviews Neuroscience},
  shortjournal = {Nat Rev Neurosci},
  volume = {8},
  number = {5},
  pages = {393--402},
  issn = {1471-003X, 1471-0048},
  doi = {10.1038/nrn2113},
  url = {https://www.nature.com/articles/nrn2113},
  urldate = {2026-04-09},
}

@article{bastos2012circuitsforpc,
  title = {Canonical microcircuits for predictive coding},
  author = {Bastos, Andre~M. and Usrey, W.~Martin and Adams, Rick~A. and Mangun, George~R. and Fries, Pascal and Friston, Karl~J.},
  year = 2012,
  month = nov,
  journal = {Neuron},
  volume = {76},
  number = {4},
  pages = {695--711},
  issn = {08966273},
  doi = {10.1016/j.neuron.2012.10.038},
  urldate = {2026-04-08},
}

@article{rao1999pc,
  title = {Predictive coding in the visual cortex: a functional interpretation of some extra-classical receptive-field effects},
  shorttitle = {Predictive Coding in the Visual Cortex},
  author = {Rao, Rajesh P. N. and Ballard, Dana H.},
  year = 1999,
  month = jan,
  journal = {Nature Neuroscience},
  volume = {2},
  number = {1},
  pages = {79--87},
  issn = {1097-6256, 1546-1726},
  doi = {10.1038/4580},
}

@article{spratling2010pc,
  title = {Predictive coding as a model of response properties in cortical area {V1}},
  author = {Spratling, Michael W.},
  year = 2010,
  month = mar,
  journal = {The Journal of Neuroscience},
  volume = {30},
  number = {9},
  pages = {3531--3543},
  issn = {0270-6474, 1529-2401},
  doi = {10.1523/JNEUROSCI.4911-09.2010},
}
%
% <OR> manually copy in the resultant .bbl file
% set second argument of \begin to the number of references
% (used to reserve space for the reference number labels box)
% \begin{thebibliography}{1}

% \bibitem{IEEEhowto:kopka}
% H.~Kopka and P.~W. Daly, \emph{A Guide to {\LaTeX}}, 3rd~ed.\hskip 1em plus
%   0.5em minus 0.4em\relax Harlow, England: Addison-Wesley, 1999.

% \end{thebibliography}

% biography section
% 
% If you have an EPS/PDF photo (graphicx package needed) extra braces are
% needed around the contents of the optional argument to biography to prevent
% the LaTeX parser from getting confused when it sees the complicated
% \includegraphics command within an optional argument. (You could create
% your own custom macro containing the \includegraphics command to make things
% simpler here.)
\begin{IEEEbiography}[{\includegraphics[width=1in,height=1.25in,clip,keepaspectratio]{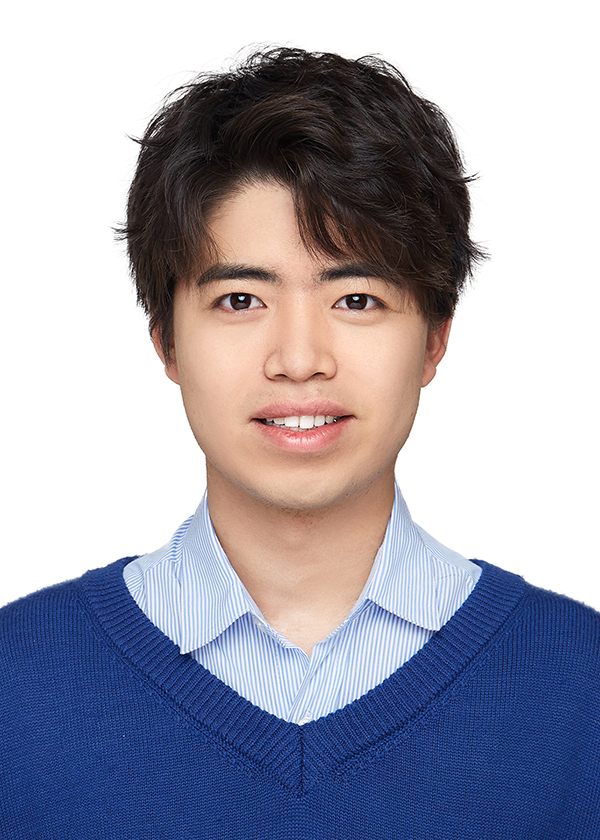}}]{Kai Li} (Student Member, IEEE) received the B.S. degree from the Department of Computer Technology and Application, Qinghai University, Xining, China, in 2020, and the M.S. degree from the Department of Computer Science and Technology, Tsinghua University, Beijing, China, in 2024, where he is currently pursuing the Ph.D. degree. His current research interests include speech/music separation, multi-modal speech separation, and audio large language models. He serves as a reviewer for several prestigious conferences and journals, including NeurIPS, ICLR, ICASSP, Interspeech, AAAI, TASLP, and TPAMI. 
\end{IEEEbiography}

\begin{IEEEbiography}[{\includegraphics[width=1in,height=1.25in,clip,keepaspectratio]{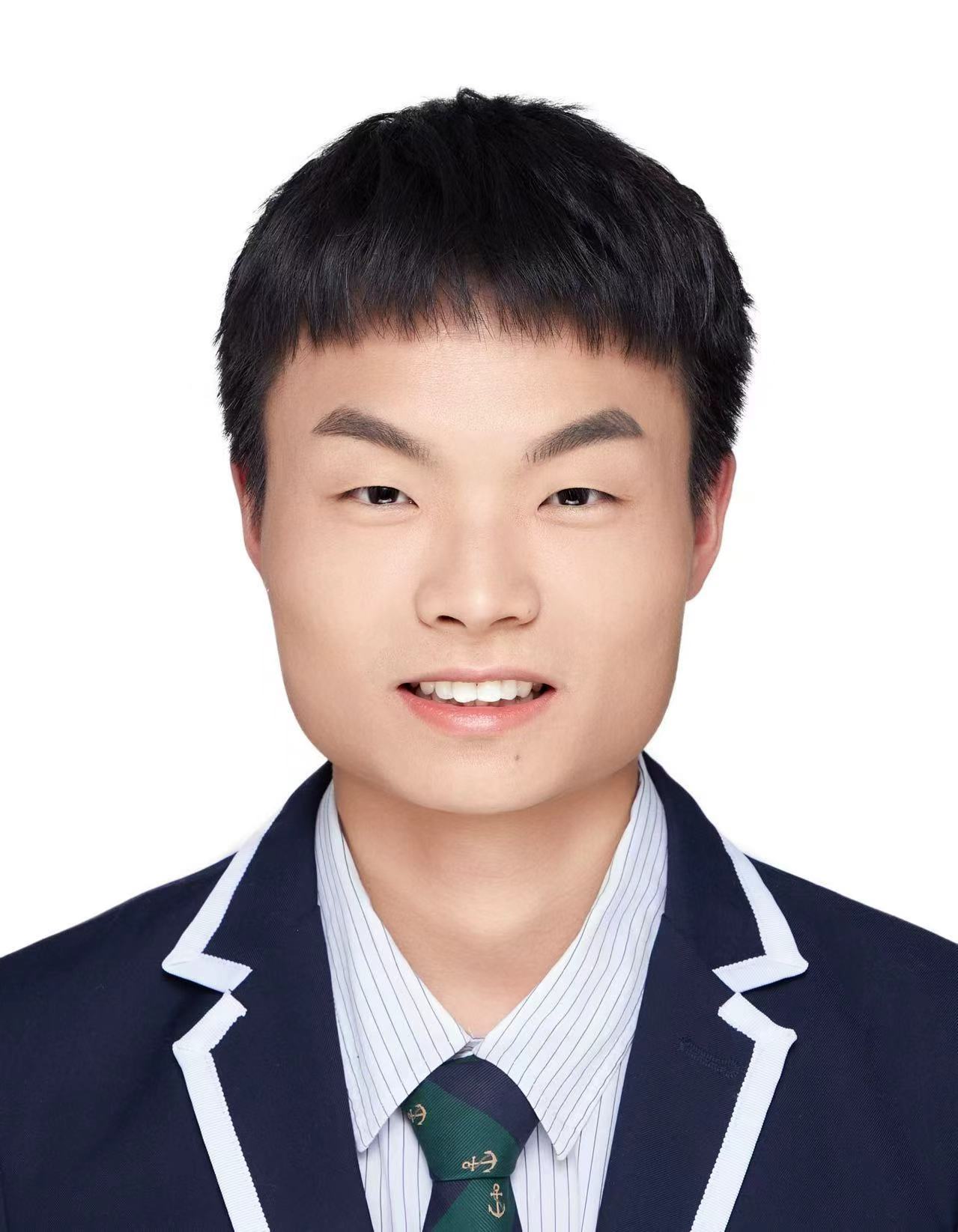}}]{Xuechao Zou} received the B.E. degree in 2021 and the M.S. degree in 2024 from the School of Computer Technology and Application at Qinghai University, Xining, China. He is currently pursuing a Ph.D. degree with the School of Computer Science and Technology at Beijing Jiaotong University. His research interests focus on computer vision, particularly remote sensing image processing.
\end{IEEEbiography}

\begin{IEEEbiography}[{\includegraphics[width=1in,height=1.25in,clip,keepaspectratio]{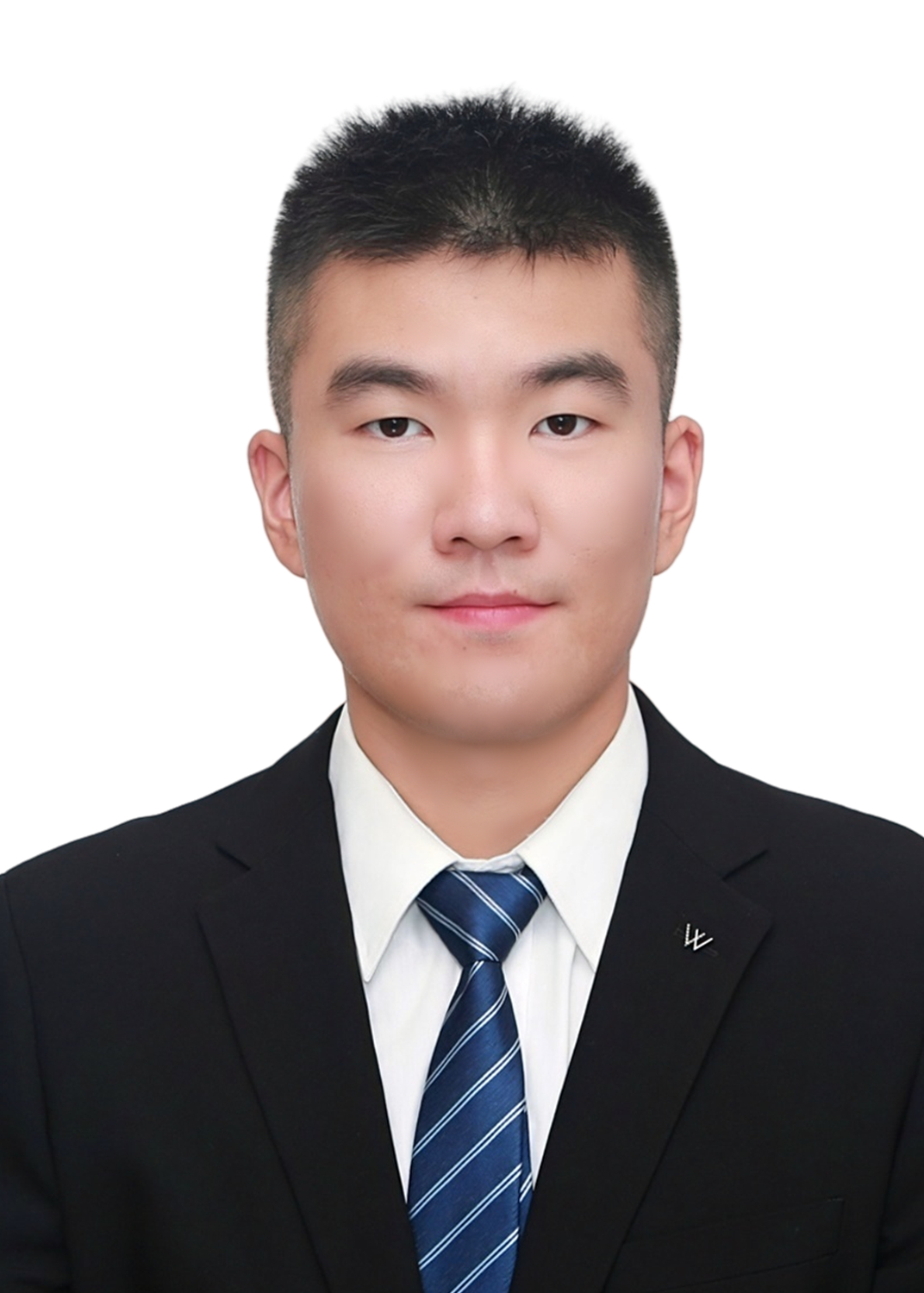}}]{Jiashen Fu} is currently pursuing the B.S. degree in the Department of Computer Science and Technology, Tsinghua University, China. His research interests include computer vision, brain-inspired artificial intelligence and computational neuroscience. 
\end{IEEEbiography}

\begin{IEEEbiography}[{\includegraphics[width=1in,height=1.25in,clip,keepaspectratio]{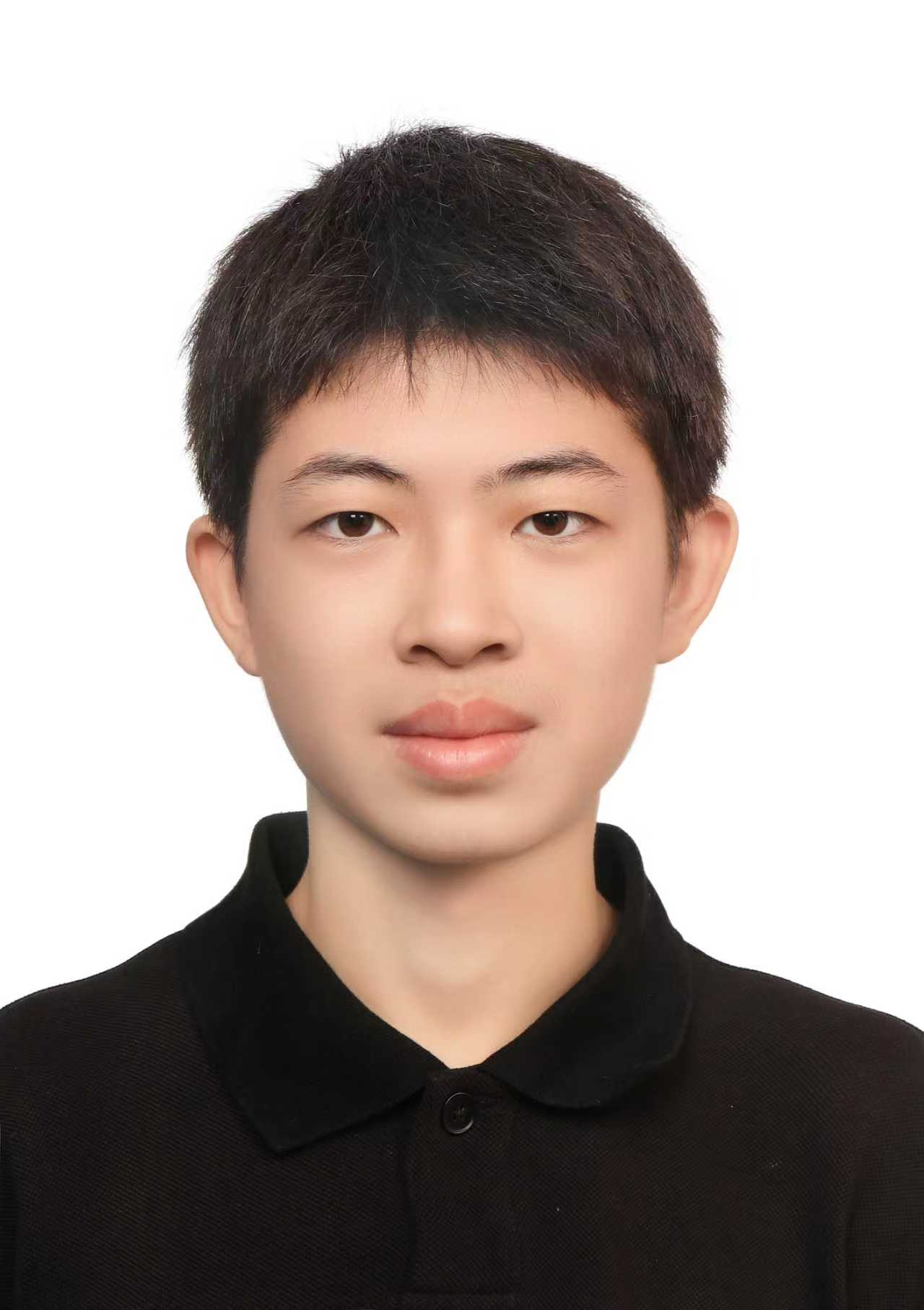}}]{Zijun Yan} is currently working toward the B.Eng. degree in Computer Science and Technology at Tsinghua University, China. His research interests include computational neuroscience.
\end{IEEEbiography}

\begin{IEEEbiography}[{\includegraphics[width=1in,height=1.25in,clip,keepaspectratio]{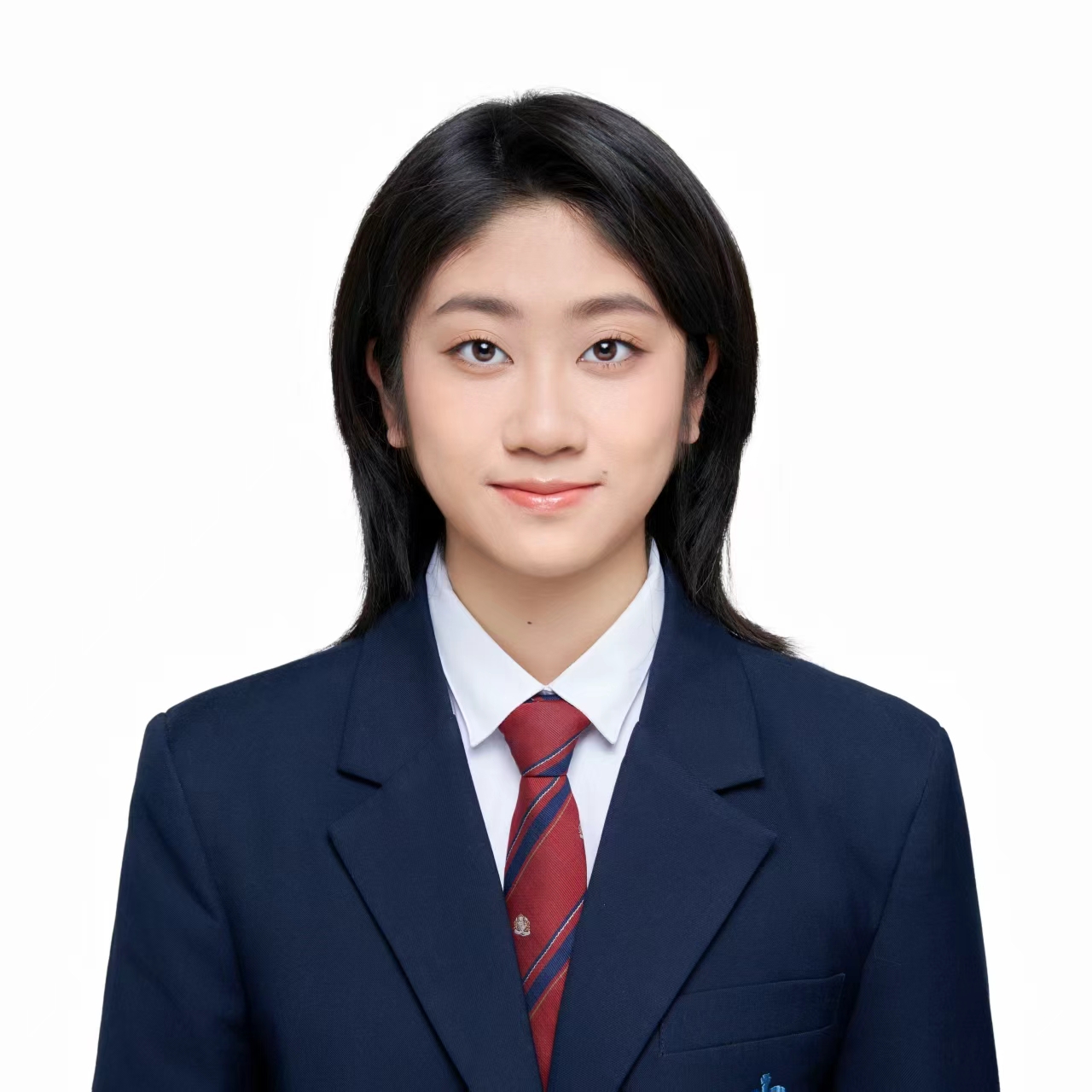}}]{Xintong Wang} is an undergraduate student in the Department of Computer Science and Technology, Tsinghua University, China (Class of 2027). Her research interests include artificial intelligence for advancing scientific discovery.
\end{IEEEbiography}

\begin{IEEEbiography}[{\includegraphics[width=1in,height=1.25in,clip,keepaspectratio]{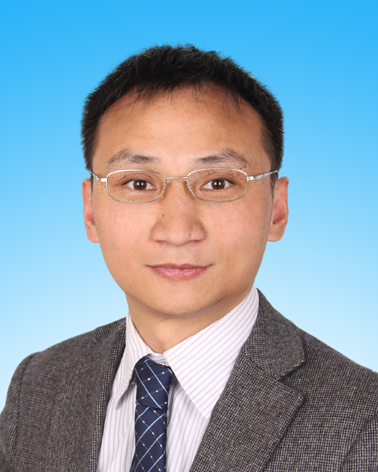}}]{Xiaolin Hu}
(S’01-M’08-SM’13) received B.E. and M.E. degrees in automotive engineering from the Wuhan University of Technology, Wuhan, China, in 2001 and 2004, respectively, and a Ph.D. degree in automation and computer-aided engineering from the Chinese University of Hong Kong, Hong Kong, in 2007. He is currently an Associate Professor at the Department of Computer Science and Technology, Tsinghua University, Beijing, China. His current research interests include deep learning and computational neuroscience. At present, he is an Associate Editor of the IEEE Transactions on Pattern Analysis and Machine Intelligence and CAAI Transactions on Intelligent Systems. Previously he was an Associate Editor of the IEEE Transactions on Image Processing, IEEE Transactions on Neural Networks and Learning Systems.
\end{IEEEbiography}
% % \begin{IEEEbiography}{Michael Shell}
% Biography text here.
% \end{IEEEbiography}

% if you will not have a photo at all:
% \begin{IEEEbiographynophoto}{John Doe}
% Biography text here.
% \end{IEEEbiographynophoto}

% insert where needed to balance the two columns on the last page with
% biographies
%\newpage

% \begin{IEEEbiographynophoto}{Jane Doe}
% Biography text here.
% \end{IEEEbiographynophoto}

% You can push biographies down or up by placing
% a \vfill before or after them. The appropriate
% use of \vfill depends on what kind of text is
% on the last page and whether or not the columns
% are being equalized.

%\vfill

% Can be used to pull up biographies so that the bottom of the last one
% is flush with the other column.
%\enlargethispage{-5in}

% that's all folks
\end{document}